\documentclass[journal]{IEEEtran}
\usepackage{amsmath,amsfonts}
\usepackage{algorithmic}
\usepackage{algorithm}
\usepackage{array}
\usepackage{textcomp}
\usepackage{stfloats}
\usepackage{url}
\usepackage{verbatim}
\usepackage{graphicx}
\usepackage{cite}
\usepackage{multirow}
\usepackage{dutchcal}
\usepackage{booktabs}
\usepackage{threeparttable}
\usepackage{subfigure}
\usepackage{color, xcolor}
\newtheorem{definition}{Definition}
\newtheorem{proposition}{Proposition}
\newtheorem{lemma}{Lemma}

\usepackage{algorithm}
\usepackage{algorithmic}

\begin{document}

\title{Interventional Causal Structure Discovery over Graphical Models with Convergence and Optimality Guarantees}

\author{Chengbo Qiu and Kai~Yang,~\IEEEmembership{Senior~Member,~IEEE}
\IEEEcompsocitemizethanks{
  \IEEEcompsocthanksitem Chengbo Qiu and Kai Yang are with the Department of Computer Science
  and Technology, Tongji University, Shanghai 201800, China, \protect\\
  E-mail: kaiyang@tongji.edu.cn \protect\\
}
}

\markboth{Journal of \LaTeX\ Class Files,~Vol.~14, No.~8, August~2021}%
{Shell \MakeLowercase{\textit{et al.}}: A Sample Article Using IEEEtran.cls for IEEE Journals}

\IEEEpubid{0000--0000/00\$00.00~\copyright~2021 IEEE}

\maketitle

\begin{abstract}
Learning causal structure from sampled data is a fundamental problem with applications in various fields, including healthcare, machine learning and artificial intelligence. Traditional methods predominantly rely on observational data, but there exist limits regarding the identifiability of causal structures with only observational data. Interventional data, on the other hand, helps establish a cause-and-effect relationship by breaking the influence of confounding variables. It remains to date under-explored to develop a mathematical framework that seamlessly integrates both observational and interventional data in causal structure learning. Furthermore, existing studies often focus on centralized approaches, necessitating the transfer of entire datasets to a single server, which lead to considerable communication overhead and heightened risks to privacy. To tackle these challenges, we develop a \textbf{b}i\textbf{l}evel p\textbf{o}lynomial \textbf{o}pti\textbf{m}ization (Bloom) framework. Bloom not only provides a powerful mathematical modeling framework, underpinned by theoretical support, for causal structure discovery from both interventional and observational data, but also aspires to an efficient causal discovery algorithm with convergence and optimality guarantees. We further extend Bloom to a distributed setting to reduce the communication overhead and mitigate data privacy risks. It is seen through experiments on both synthetic and real-world datasets that Bloom markedly surpasses other leading learning algorithms. 
\end{abstract}


\section{Introduction}
\IEEEPARstart{C}{ausal} structure learning aims to learn the directed acyclic graph (DAG) of causal graphical models from sampled data, which enables us to reveal and understand the potential causal relationships among different variables \cite{pearl2000models}. Recently, it has not only emerged in various artificial intelligence tasks, such as Natural Language Processing \cite{niu2021counterfactual}, Reinforcement Learning \cite{yang2023reinforcement}, and Anomaly Detection \cite{lin2022causal}, but also played an essential role in other domains, such as healthcare \cite{peyrot1996causal}, economics, and geosciences \cite{prill2010towards}. Learning DAGs from data, however, is regarded as an NP-hard problem \cite{chickering1996learning}, mainly owing to the acyclic constraints. Traditional methods, such as constraint-based \cite{spirtes2000causation, runge2019detecting, runge2018conditional}, score-based methods \cite{chickering2002optimal} and functional causal models (FCMs) \cite{shimizu2006linear}, typically search for causal graphs in a discrete manner based on some assumptions on the data and the underlying mechanisms. However, the large search space of DAGs makes these methods suffer from computational inefficiency. Zheng et al. \cite{zheng2018dags} proposed Notears, which describes acyclic graphs with a smooth function over real matrices, and formulates the causal structure learning problem as a constrained optimization problem that can be successively optimized. And this work provides a foundation for subsequent research \cite{yu2019dag, zhu2019causal}. 
\IEEEpubidadjcol

However, these works \cite{yu2019dag,ng2022masked} often employ gradient descent-type algorithms such as SGD, which may lead to being stuck in locally optimal solutions and saddle points or experiencing gradient explosion or oscillations \cite{ng2024structure}. These issues often cause slow or unstable convergence, as discussed in \cite{hanin2018neural}. Moreover, the gradient descent method is sensitive to data noise and outliers. In contrast, many global optimization methods, such as polynomial optimization (POP) methods, demonstrates superior convergence and robustness, and provide the potential for globally optimal solutions under mild assumptions \cite{lasserre2001global}. Secondly, previous algorithms primarily rely on observational data, which poses theoretical limits in identifying true DAGs \cite{yang2018characterizing}. Interventions are now extensively applied in various real-world contexts, including genomics and microservice systems \cite{lin2022causal}. The introducing of interventional data can efficiently improve the identifiability of causal structures \cite{eberhardt2012almost}, and help establish causal relationships. Although some current research \cite{wang2017permutation,lorch2022amortized} on causal structure learning has already incorporated interventional data, these studies either present high complexity issues \cite{wang2017permutation, squires2020permutation}, or need a more generalized framework to integrate both types of data through a unified optimization strategy \cite{lippe2021efficient}. Also, many existing works typically lack sufficient theoretical foundations including guarantees of convergence and optimality. Consequently, developing a mathematical framework for causal structure learning that not only provides theoretic support but also integrates observational and interventional data remains a significant challenge. Lastly, many existing studies generally focus on centralized approaches, which may result in substantial communication overhead and higher privacy risks. When compared with some current federated-learning based works \cite{huang2023towards, ng2022towards}, our approach can efficiently reduce the identification of spurious causalities by incorporating interventional data. \cite{abyaneh2022fed} is also a distributed algorithm that uses interventional data, but it usually requires a sufficient amount of data to train a local neural network model on each client, and they lack convergence and optimality guarantees.

To this end, we propose an algorithm for causal structure learning with bilevel polynomial optimization (Bloom), which offers a robust mathematical modeling approach with theoretical support for learning causal structure from both interventional and observational data. And it can efficiently address the causal structure learning issue with convergence and optimality guarantees by solving a series of semidefinite (SDP) relaxation problems. We further expand Bloom to a distributed setting to reduce communication overhead and mitigate data privacy risks.

Our contributions can be summarized as follows:
\begin{enumerate}
\item We propose a bilevel polynomial optimization modelling framework for causal structure discovery from both observational and interventional data. This framework allows us not only to seamlessly integrate the observational data with interventional data in the causal structure learning process, but also offers a fundamental yet unique perspective to the continuous optimization problem associated with the search of an optimal DAG of causal graph model. Given the fact that there exist a large body of efficient algorithms for bilevel and polynomial optimization problem, the proposed bilevel framework opens up new avenues for modelling and analyzing causal structure learning from both observational and interventional data.
\item Existing continuous optimization-based algorithms often use gradient descent-type methods e.g., SGD to discover high-scoring causal structures. Such methods may get stuck at local optima or saddle points, which are notoriously difficult for SGD to escape. Instead, building upon the proposed bilevel polynomial optimization model, we delve into its unique structure and theoretically demonstrate its convertibility into a single-level optimization problem. Leveraging this reformulation, we introduce the Bloom algorithm, which can gradually approximate the global optimal solution for the causal structure learning problem and is capable of escaping local optima or saddle points in the searching process.
\item Most works in the literature focus on centralized methods for causal structure learning, which can result in issues like high communication overhead, inadequate computation power, and potential data privacy breaches. Therefore, we further extend the proposed algorithm into distributed systems. The proposed algorithm does not require sharing the client's local data, but only the learned model parameters. This ensures privacy protection requirements and has lower communication pressure.
\end{enumerate}

\section{Background and Related works}
\subsection{Causal Structure Learning}

Causal structure learning is defined as learning a DAG, represented as $\mathbb{G}(V,E)$, over the data $\boldsymbol{X} \in \mathbb{R}^{N \times D}$ sampled from a joint distribution $\mathbb{P}(X)$. In graph $\mathbb{G}$, each node $i\in V$ corresponds to a random variable $X_i \in \left\{X_1, \ldots, X_D\right\}$, and edge $(i,j)\in E$ denotes a direct causal relationship from the variable $X_i$ to $X_j$, i.e., $X_i \rightarrow X_j$. The distribution of variable $X_i$ is $\mathbb{P}_i\left(X_i \mid \mathrm{Pa}\left(X_i\right)\right)$, where $\mathrm{Pa}\left(X_i\right)$ denotes the parent set of node $X_i$. Intervention on variable $X_i$ is defined as replacing the conditional distribution $\mathbb{P}\left(X_i \mid \mathrm{Pa}\left(X_i\right)\right)$ with a new distribution $\mathbb{P}_\mathrm{new}$, including perfect and imperfect interventions. Perfect intervention refers to removing the effects of all parent variables, i.e., $\mathbb{P}_\mathrm{new}= \widetilde{\mathbb{P}}(X_i )$; while imperfect intervention replaces the original conditional distribution with a new conditional distribution, i.e., $\mathbb{P}_\mathrm{new}= \widetilde{\mathbb{P}}(X_i |\mathrm{Pa}(X_i ))$.

There are many causal structure learning methods, which can be broadly categorized into FCMs, constraint-based and score-based methods. FCMs aim to learn and represent the causal relationship between cause and effect variables using a predefined function containing independent noise terms. Typical FCMs include LiNGAM \cite{bib37}, PNL \cite{bib38} and ANM \cite{bib39}. Constraint-based methods leverage conditional independence (CI) tests between the variables to identify causal relationships, such as PC and FCI\cite{spirtes2000causation}. COmbINE \cite{triantafillou2015constraint} and HEJ \cite{hyttinen2014constraint} support the introduction of interventional data, and they typically rely on Boolean satisfiability solvers. Mooij et al. \cite{mooij2020joint} proposed the joint causal inference framework that can handle unknown interventions with by coupling with many constraint-based algorithms. Score-based methods aim to obtain a DAG by optimizing a score function. GES \cite{chickering2002optimal} searches for the highest scoring graphs from a discrete space $\mathbb{G}$ by iteratively adding, removing and flipping edges. GIES \cite{hauser2012characterization} and GNIES \cite{gamella2022characterization} are the variants of GES that can be used for interventional data. IGSP is a hybrid method \cite{squires2020permutation}. However, these methods typically search in discrete spaces.

\paragraph{Causal Structure Learning with Continuous Optimization} Zheng et al. \cite{zheng2018dags} proposed Notears, which utilizes a smooth function $\mathcal{h}(\boldsymbol{W})$ over real matrices to encode the acyclic constraint, and transformed the aforementioned problem into the following continuous optimization problem:
\begin{equation}
\begin{array}{ll}\min & \mathcal{S}(\boldsymbol{W}) \\ \text { s.t. } & \mathcal{h}(\boldsymbol{W})=\operatorname{tr}\left(e^{\boldsymbol{W} \odot \boldsymbol{W}}\right)-d=0 \\ \text { var } & \boldsymbol{W}{.}\end{array}
    \label{e1}
\end{equation}

The function $\mathcal{h}(\boldsymbol{W})=0$ holds if and only if graph $\mathbb{G}$ corresponding to $\boldsymbol{W}$ is a DAG. Let $\boldsymbol{W}'=\boldsymbol{W} \odot \boldsymbol{W}$, where $\odot$ denotes the Hadamard product. And the acyclic constraint is equivalent to $\sum_{k=1}^{\infty} \alpha_k \sum_{i=1}^D({\boldsymbol{W}^{\prime}}^k)_{i i}=0$ by performing a Taylor expansion of $\mathcal{h}(\boldsymbol{W}^{\prime})$, with $\alpha_k$ denoting weight coefficients. And $\sum_{i=1}^D({\boldsymbol{W}^{\prime}})^k_{i i}$ counts the sum of length-k weighted closed walks in directed graphs. Notears uses linear SEM as the causal model and sets the scoring function $\mathcal{S}$ to a least squares loss. Based on Notears, many subsequent studies and improvements have been developed, e.g., DAG-GNN \cite{yu2019dag}, and Gran-DAG \cite{lachapelle2019gradient}. Deng et al. \cite{deng2023optimizing} proposed a bilevel optimization algorithm, in which it defines a constraint set based on KKT conditions to guide the search of topological order. But it can only iteratively search for local minimum and lacks the guarantee of optimality of the results. Wei et al. \cite{wei2020dags} generalized existing acyclic constraints Eq. (\ref{e1}) to a class of matrix polynomials and defined an explicit edge absence constraint set. It works by iteratively adding and removing elements to the constraint set until it reaches what they call "irreducibility". However, it can only guarantee local optimal solutions. Ng et al. \cite{ng2022convergence} studied the optimality conditions and convergence property of the augmented Lagrangian method (ALM) and the quadratic penalty method (QPM) in structural learning problems, and Deng et al. \cite{deng2024global} showed that a proper optimization method converges to the global minimum of the least squares objective in the bivariate case. Ng et al. \cite{ng2024structure} investigated the performance of continuous structure learning methods under different noise variances and analyzed the possible reasons.

\paragraph{Causal Structure Learning with Interventional Data} Interventional Greedy SP (IGSP) algorithm \cite{wang2017permutation} is a hybrid approach that uses conditional independence tests in the score function and can be used with interventional data. Brouillard \cite{brouillard2020differentiable} introduced a novel differentiable causal structure learning method, named DCDI, leveraging neural networks, designed to harness the potential of interventional data. Meanwhile, ENCO \cite{lippe2021efficient} concentrated on both observational and interventional data, employing a methodology involving two alternating processes: fitting distributions and learning the causal structure. ENCO also uses two variables to parameterize the graph structure. While these methods prove effective, their reliance on gradient descent-based methods introduces the possibility of converging to local optima \cite{brouillard2020differentiable, lippe2021efficient}. This emphasizes the need for further exploration into causal structure learning methods with global search. 

\paragraph{Causal Structure Learning with Distributed Learning}
However, these methods usually require centralized data, which can easily cause privacy leakage and high communication consumption problems when these data come from different data sources. To this end, there are some works that study the problem of causal structure learning in distributed environments. Huang et al. \cite{huang2023towards} proposed a federated PC algorithm, which designed a layer-wise strategy to identify consistent separation sets among clients and identify accurate edge orientation without centralizing data from each client to the server. Li et al. \cite{li2024federated} proposed a federated constraint-based method for heterogeneous data, which protects data privacy by aggregating statistics of the raw data on different clients. Abyaneh et al. \cite{abyaneh2022fed} developed a federated framework (FedCDI) for inferring causal structures from distributed data containing interventional samples, which can uncover the underlying causal structure by exchanging belief updates of the clients without sharing local samples. Ng et al. \cite{ng2022towards} develop a distributed structure learning method based on continuous optimization, using the alternating direction method of multipliers (ADMM). FedDAG \cite{gao2021feddag} is also a continuously optimized federated structure learning algorithm, which includes aggregation of graph structures and approximation of local mechanisms to accommodate the data heterogeneity of clients. But most of them only use observational data, which may lead to the spurious causal relationships.

\begin{table}[htb]
\caption{Notations\label{tab:table1}}
\renewcommand\arraystretch{1.1}
\resizebox{\linewidth}{!}{
\begin{tabular}{lll}
\specialrule{0em}{1.5pt}{1.5pt}
\midrule
\multicolumn{2}{l}{\textbf{Symbol}}         & \textbf{Description}           \\ \hline
\multirow{4}{*}{Universal symbol} & $X$       & Variable                             \\
                                  & $\boldsymbol{M}$        & Matirx                             \\
                                  & $\boldsymbol{m}$        & Vector                             \\
                                  & $\mathcal{F}/\mathcal{f}$        & Function                             \\
                                  & $\left[D\right]$ & Set, \{1,...D\}                     \\ \hline
\multirow{9}{*}{Special symbol}   & $\mathbb{G}$        & Graph                              \\            & $V$        & Nodes of the graph 
                                \\& $E$        & Edges of the graph \\
                                  & $\mathbb{P}$        & Probability distribution                           \\
                                  & $\mathbb{S}$       & Symmetric matrix             \\
                                  & $\mathbb{S}_{+}$        & Positive semidefinite matrix                               \\
                                  & $\mathbb{D}$        & DAG space \\ 
                                  & $\boldsymbol{x}^{\boldsymbol{\alpha}}$  & Monomial \\
                                  & $\boldsymbol{x}^{\mathbb{N}_{d}^N}/ \mathbb{N}_{d}^N$ & Standard monomial basis \\
                                  \midrule
\specialrule{0em}{1.5pt}{1.5pt}
\end{tabular}}
\end{table}

\subsection{Bilevel Optimization}

A bilevel optimization problem is an optimization problem with two levels, each of which has its own objective function and constraints. The general form can be denoted as:
\begin{equation}
\begin{array}{ll}\min & \mathcal{F}(\boldsymbol{x}, \boldsymbol{y}) \\ \text { s.t. } & \mathcal{f}_i(\boldsymbol{x}, \boldsymbol{y}) \leq 0, \quad i=1, \ldots, I \\ & \boldsymbol{y}=\operatorname{argmin}_{\boldsymbol{y}^{\prime}} \mathcal{G}\left(\boldsymbol{x}, \boldsymbol{y}^{\prime}\right) \\ & \text { s.t. } \mathcal{g}_j(\boldsymbol{x}, \boldsymbol{y}) \leq 0, \quad j=, \ldots, J \\ \text { var } & \boldsymbol{x}, \boldsymbol{y}{,}\end{array}
    \label{e2}
\end{equation}
where $\mathcal{F}$ and $\mathcal{G}$ are the upper- and lower-level objective function, $\boldsymbol{x}$ and $\boldsymbol{y}$ are decision variables, $\mathcal{f}_i$ is the $i_{th}$ upper-level constraint, and $\mathcal{g}_j$ denotes the $j_{th}$ lower-level constraint.

There are many approaches for bilevel optimization \cite{gould2016differentiating}, here we focus on the method based on optimal conditions. This method aims to replace the lower-level optimization problem with the optimal conditions (e.g., Karush-Kuhn-Tucker (KKT) conditions) \cite{biswas2019literature} and reformulates it as a Mathematical Problem with Complimentary Constraints, which can be expressed as follows.
\begin{equation}
\begin{array}{ll}\min & \mathcal{F}(\boldsymbol{x}, \boldsymbol{y}) \\ \text { s.t. } & \mathcal{f}_i(\boldsymbol{x}) \leq 0, \quad i=1, \ldots, I \\ & \mathcal{g}_j(\boldsymbol{y}) \leq 0, \quad j=1, \ldots, J \\ & \nabla_{\boldsymbol{y}} \mathcal{G}(\boldsymbol{x}, \boldsymbol{y})+\sum_{j=1}^M \lambda_j \nabla_{\boldsymbol{y}} \mathcal{g}_j(\boldsymbol{x}, \boldsymbol{y})=0 \\ & \lambda_j \mathcal{g}_j(\boldsymbol{x}, \boldsymbol{y})=0, \quad j=1, \ldots, M \\ & \lambda_j \geq 0 \\ \operatorname{var} & \boldsymbol{x}, \boldsymbol{y}{.}\end{array}
    \label{e3}
\end{equation}

Moreover, the problem is shown to be equivalent to the original bilevel optimization problem when Slater's condition (\textit{Definition 1}) is satisfied \cite{dempe2012bilevel}.

\begin{definition}[Slater’s Condition]
Slater’s constraint qualification for the lower-level problem at any parameter $\boldsymbol{x}$: there exist $\boldsymbol{y}^*$ such that $\mathcal{g}_j(\boldsymbol{x}, \boldsymbol{y}^*)<0$, $j=1, \ldots, M$.
\end{definition}

\subsection{Polynomial Optimization}

The general form of the polynomial optimization problem (POP) denotes as follows:
\begin{equation}
    \begin{array}{ll}\min & \mathcal{F}(\boldsymbol{x}) \\ \text { s.t. } & \mathcal{f}_i(\boldsymbol{x}) \geq 0, \quad i=1, \ldots, I \\ \operatorname{var} & \boldsymbol{x}{.} \end{array}
    \label{e4}
\end{equation}

The objective function $\mathcal{F}$ and constraints $\mathcal{f}_i$ are assumed to be polynomials in the variables $\boldsymbol{x}=(x_1,\dots,x_N)$, expressed as $\mathcal{f}(\boldsymbol{x})=\sum_{\boldsymbol{\alpha} \in A} f_\alpha \boldsymbol{x}^{\alpha}$, where $f_\alpha\in\mathbb{R}$ represents the coefficients, and $A\subseteq\mathbb{N}^N$. The monomial $\boldsymbol{x}^{\boldsymbol{\alpha}}$ denotes $x_1^{\alpha_1} \dots x_N^{\alpha_N}$, where $\boldsymbol{\alpha}=(\alpha_1,\dots,\alpha_N )$ is the order of each variable $x_n$. The degree of monomial $\boldsymbol{x}^{\boldsymbol{\alpha}}$ is $\operatorname{deg}(\boldsymbol{x}^{\boldsymbol{\alpha}})=\sum_{i=1}^N \alpha_i$. The support set of a polynomial function $\mathcal{f}$ is defined as $\operatorname{supp}(\mathcal{f}):=\left\{\boldsymbol{\alpha} \in A \mid f_\alpha \neq 0\right\}$, and the degree of a polynomial function $\mathcal{f}$ is $\operatorname{deg}(\mathcal{f})=\max \{\operatorname{deg}(\boldsymbol{x}^{\boldsymbol{\alpha}}): \boldsymbol{\alpha} \in \operatorname{supp}(\mathcal{f})\}$. Let $\mathbb{R}\left[x\right]$ be the set of real $n$-variate polynomials, and the set of polynomials of degree no more than $2d$ is represented as $\mathbb{R}_{2d} [x]$. For $r\in \mathbb{Z}_{+}$, let $\mathbb{S}^{r\times r}$ (resp. $\mathbb{S}_{+}^{r\times r}$) denote the set of symmetric matrices (resp. positive semidefinite matrices).

Traditional approaches \cite{lasserre2001global} achieve solutions by constructing a sum-of-squares (SOS) hierarchy for Eq. (\ref{e4}) and solving a sequence of corresponding SDP problems. The polynomial is expressed as $f(\boldsymbol{x})=(\boldsymbol{x}^{\mathbb{N}_{d}^N})^T \boldsymbol{Q}(\boldsymbol{x}^{\mathbb{N}_{d}^N} )$, where $\boldsymbol{Q}\in \mathbb{S}_{+}^{r\times r}$ is the Gram Matrix \cite{reznick1978extremal}. Here, $\boldsymbol{x}^{\mathbb{N}_{d}^N}$ represents the standard monomial basis of $\boldsymbol{x}$ with a degree not exceeding $d$, \textit{denoted as $\mathbb{N}_{d}^N$ in the sequel}. Let $\boldsymbol{y}$ be the univariate vector corresponding to $\mathbb{N}{d}^N$, then the monomial variable $\boldsymbol{x}^{\boldsymbol{\alpha}}$ can be replaced by a real variable $y_{\boldsymbol{\alpha}}$. Define $\mathcal{L}_{\boldsymbol{y}}:\mathbb{R}\left[\boldsymbol{x}\right]\rightarrow \mathbb{R}$ as the linear function:
\begin{equation}
\mathcal{f}(\boldsymbol{x})=\sum_{{\boldsymbol{\alpha}} \in A} f_{\boldsymbol{\alpha}} \boldsymbol{x}^{\boldsymbol{\alpha}} \rightarrow \mathcal{L}_{\boldsymbol{y}}(\mathcal{f})=\sum_{{\boldsymbol{\alpha}} \in A} f_{\boldsymbol{\alpha}} y_{\boldsymbol{\alpha}}{.}
    \label{e5}
\end{equation}

Let $\boldsymbol{M}_d (\boldsymbol{y})$ be the $d$-order moment matrix associated with $\boldsymbol{y}$. The items in the matrix can be obtained as follows:
\begin{equation}
\boldsymbol{M}_d(\boldsymbol{y})_{a b}=\mathcal{L}_{\boldsymbol{y}}\left(\mathbb{N}_d^N[a] \mathbb{N}_d^N[b]\right)=y_a y_b{,} 
    \label{e6}
\end{equation}
where $1\leq a,b \leq |\boldsymbol{y}|$ denote indexes, and $|\boldsymbol{y}|$ denotes the dimension of $\boldsymbol{y}$.

For the constraint function $\mathcal{f}_i(\boldsymbol{x})$, the elements in the $d$-order localizing matrix $\boldsymbol{M}_d (\mathcal{f}_i\boldsymbol{y})$ are presented as:
\begin{equation}
    \boldsymbol{M}_{d}(\mathcal{f}_i\boldsymbol{y})_{ab}=\mathcal{L}_{\boldsymbol{y}}(\mathcal{f}_i\mathbb{N}_{d}^{N}[a]\mathbb{N}_{d}^{N}[b])=\sum_{\boldsymbol{\alpha}\in A}y_{\boldsymbol{\alpha}}y_{a}y_{b}{,}
    \label{e7}
\end{equation}
where $y_{\alpha}=\mathcal{L}_{y}(\boldsymbol{x}^{\alpha})$. Let $d_i=\lceil \deg(\mathcal{f}_i)/2\rceil (i=1,...,I)$ and $d_{\mathrm{min}}=\max\{\lceil\mathrm{deg}(\mathcal{F})\rceil/2,d_1,...,d_l\}$.

By introducing Eq. (\ref{e6}) and (\ref{e7}), the problem Eq. (\ref{e4}) can be relaxed into a SDP problem (with relaxation order $d\geq d_{\mathrm{min}}$):
\begin{equation}
\begin{array}
{ll}\min & \mathcal{L}_{\mathbf{y}}(\mathcal{F})  \\ \text { s.t. }& \boldsymbol{M}_{d}(\boldsymbol{y})\in\mathbb{S}_{+}  \\
&\boldsymbol{M}_{d-d_{i}}(f_{i}\boldsymbol{y})\in\mathbb{S}_{+}, \quad i=1,\ldots,I  \\
\operatorname{var}& \boldsymbol{y}{.} \\
\end{array}
    \label{e8}
\end{equation}

However, it is more complicated to solve the above problem directly due to the scale of the issue \cite{grone1984positive}.

\subsubsection{Correlative sparsity} H. Waki \cite{waki2006sums} proposed a sparse SOS hierarchy by utilizing the correlative sparsity among variables. The method defines the correlative sparsity pattern (CSP) graph associated with Eq. (\ref{e4}) as $\mathbb{G}^{\mathrm{csp}}(V,E)$. If nodes $a$, $b\in V$ both appear in a constraint or a monomial, then $(a,b)\in E$.

According to definitions of chordal graph (\textit{Definition 2}) and clique (\textit{Definition 3}), the graph $\mathbb{G}^{\mathrm{csp}}$ can be extended to a chordal graph ${(\mathbb{G}^{\mathrm{csp}})}'$ by adding appropriate edges. Thus, the graph ${(\mathbb{G}^{\mathrm{csp}})}'$ can be readily divided in cliques (\textit{Theorem 2.3} in \cite{agler1988positive}). Let $\mathbb{C}=\{C_{l}\}_{l=1}^{L}$ represent the set of maximal cliques in ${(\mathbb{G}^{\mathrm{csp}})}'$, the variables in the clique $C_l$ are denoted as $\boldsymbol{x}[C_{l}]$. And the constraint polynomials $\mathcal{f}_{1},\ldots,\mathcal{f}_{I}$ can be grouped into sets $\mathbb{J}=\{J_{l}\}_{l=1}^{L}$ according to whether the variables in $\mathcal{f}_{i}\in J_{l}$ belong to clique $C_l$. The sets $J_1,\dots,J_L$ are mutually exclusive, and $\bigcup_{l=1}^{L}J_{l}=[1,\dots,I]$.

\begin{definition}[Chord and Chordal Graph]
A chord is defined as an edge between two nonconsecutive nodes in a cycle. A graph is said to be called a chordal graph if all cycles of length at least 4 have a chord.
\end{definition}

\begin{definition}[Clique and Maximal Clique]
A complete graph is a graph where every pair of nodes is connected by an edge. A clique in a graph is a group of nodes that form a complete subgraph. A maximal clique is a clique that is not a subset of any other clique.
\end{definition}

Based on the cliques, the SDP matrices can be separated into blocks. And the relaxation problem based on correlative sparsity for the POP Eq. (\ref{e4}) is denoted as:
\begin{equation}
\resizebox{.9\linewidth}{!}{$
            \displaystyle
\begin{array}
{ll}\min & \mathcal{L}_{\mathbf{y}}(\mathcal{F})  \\
\text{s.t.} &  \boldsymbol{M}_{d}(\boldsymbol{y},C_{l})\in\mathbb{S}_{+},\quad l=1,\ldots,L  \\
&\boldsymbol{M}_{d-d_{j}}(\mathcal{f}_{j}\boldsymbol{y},C_{l})\in\mathbb{S}_{+},\quad j\in J_{l},l=1,\ldots,L \\
\mathrm{var~}&  \boldsymbol{y}{.}
\end{array}$}
    \label{e9}
\end{equation}

\subsubsection{Term sparsity} Wang \cite{wang2021tssos} proposed a sparse SOS with term sparsity by using correlations between monomials (terms). This method constructs a term sparsity pattern (TSP) graph, denoted as $\mathbb{G}^{\mathrm{tsp}}(V,E)$. The nodes in $V$ correspond to the terms $\mathbb{N}_d^N$ in the monomial basis, and the set of edges is $E=\left\{\{\beta,\gamma\}|\beta\neq\gamma\in V,\beta+\gamma\in A\cup(2\mathbb{N})^N\right\}$, where $(2\mathbb{N})^N:=\{2\boldsymbol{\alpha}|\boldsymbol{\alpha}\in\mathbb{N}^N\}$ and $A=\operatorname{supp}(\mathcal{F})\cup\cup_{i=1}^{I}\operatorname{supp}(\mathcal{f}_{i})$.

The TSP subgraph for the constraint polynomial $\mathcal{f}_i$ is represented as $\mathbb{G}_{d,i}^{\mathrm{tsp}}(V_{d,i},E_{d,i})$, where $V_{d,i}$ corresponds to the monomials in $\mathbb{N}_{d-d_{i}}^{N}$. The edges $E_{d,i}$ in the graph are constructed through the iterative process, involving two successive operations: support extension and chordal extension.

With $d \geq d_{\mathrm{min}}$, the relaxation problem based on term sparsity for the POP Eq. (\ref{e4}) is defined as:
\begin{equation}
\resizebox{.92\linewidth}{!}{$
            \displaystyle
\begin{array}
{ll}\min& \mathcal{L}_{\mathbf{y}}(\mathcal{F})  \\
\text{s.t.}& B_{d}^{\mathrm{tsp}}\odot M_{d}(\mathbf{y})\in\prod_{\mathbb{G}_{d}^{\mathrm{tsp}}}(\mathbb{S}_{+})  \\
&B_{d,i}^{\mathrm{tsp}}\odot M_{d-d_{i}}(\mathcal{f}_{i}\boldsymbol{y})\in\prod_{\mathbb{G}_{d,i}^{\mathrm{tsp}}}(\mathbb{S}_{+}),\quad i=1,\ldots,I \\
\text{var}&  \boldsymbol{y}{,}
\end{array}$}
    \label{e10}
\end{equation}
where $B_{d}^{\mathrm{tsp}}$ represents the adjacency matrix of $\mathbb{G}_{d}^{\mathrm{tsp}}$ and $\boldsymbol{M}\in\Pi_{\mathbb{G}_{d}^{\mathrm{tsp}}}(\mathbb{S}_{+})$ denotes that the principal submatrices in the matrix $\boldsymbol{M}$ are all positive semidefinite.

\begin{figure*}[htb]
\centering
\includegraphics[width=1.7\columnwidth]{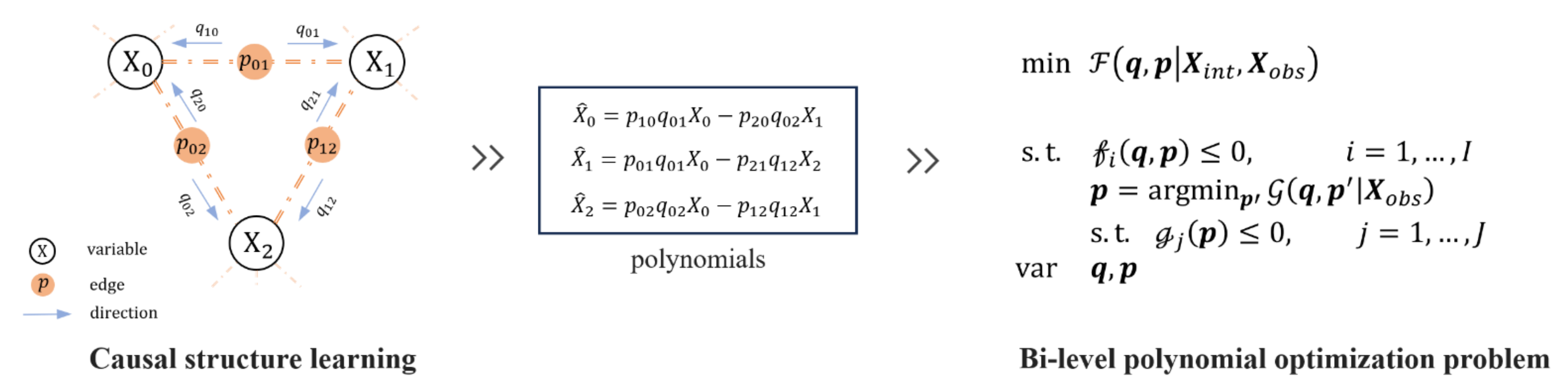}
\caption{Framework of the proposed algorithm.}
\label{fig_1}
\end{figure*}

\section{Methodology}
In this section, we first discuss the background of using Bloom for causal structure learning given both observational and interventional data. Then we present our method in detail, as illustrated in Fig. \ref{fig_1}, and discuss the convergence guarantees. Finally, we introduce the distributed architecture of Bloom.

\subsection{Scope and Assumptions}
We aim to learn a DAG of an causal graphical model given observational and intervention data. Here, we first make the causal faithful and causal sufficient assumptions for the causal model, i.e., the observed data is consistent with the real causal relationship, and all variables for inferring the causal relationship are included. However, in the following, we relax this assumption appropriately and experimentally discuss the causal structure discovery performance of the proposed method in scenarios with potentially confounding causal variables.

Furthermore, we assume that interventions are created by affecting only a single variable, and each intervention is independent of the other. The scope of this intervention closely follows Ke et al. \cite{ke2019learning}. Besides, we assume that all interventions are perfect, i.e., the distribution of the intervened variable is independent of the parent variables. And we assume interventions are conducted on each variable, and all samples, including the intervention target, are provided. In next section, we also try to relax this assumption and experimentally discuss the performance of the proposed method in the case of imperfect intervention.

\subsection{Bilevel Polynomial Optimization for Causal Structure Learning}

Pearl's Causal Hierarchy theory \cite{pearl2000models} posits that causal inference consists of three ascending levels: association, intervention, and counterfactual. Each level cannot provide higher-level information. In the first level, association problem is typically addressed by using passively observed data. On the second level, interventional data can offer more information of directed causality. However, learning causal structures relying solely on observational data is typically challenging. This is because, under the faithfulness assumption, the model may only be able to learn the Markov equivalence class of the true graph, whereas interventional data can effectively enhance identifiability. 

Therefore, we can define two parameters to jointly determine the graph structure, respectively: $\boldsymbol{P}\in[-1,1]^{D\times D}$ denotes the existence of undirected edges, $\boldsymbol{Q}\in[0,1]^{D\times D}$ the direction of the edges. In particular, the diagonal entries of both $\boldsymbol{P}$ and $\boldsymbol{Q}$ are $0$, i.e., $P_{ii}$ (or $Q_{ii}$ )=$0$, $i\in[D]{:=}\{1,\ldots,D\}$, besides $P_{ij}=P_{ji}$, $Q_{ij}+Q_{ij}=1$. Then the weighted adjacency matrix of the DAG can be determined by $\boldsymbol{W}=\boldsymbol{P}\odot \boldsymbol{Q}$. When $W_{ij}=P_{ij}\cdot Q_{ij}=0$, it means there is no directed causal relationship between variable $i$ and $j$. By decomposing the adjacency matrix into these two parameters, we can model the causal structure learning with observational and interventional data as a bilevel optimization problem thereby improving the identifiability of the structure. However, this modelling may entail extensive matrix operations, particularly with acyclic constraints, resulting in heightened complexity and slow convergence. Let $\boldsymbol{p}\in[-1,1]^{D_{2}}$ and $\boldsymbol{q}\in[0,1]^{D_{1}}$ be the concatenation of the non-diagonal elements from the upper triangular matrices of the matrices $\boldsymbol{P}$ and $\boldsymbol{Q}$, respectively, where $\overline{D}=D_{1}=D_{2}=\frac{D^{2}-D}{2}$.Thus, the elements in $\boldsymbol{P}$ and $\boldsymbol{Q}$ can be easily expressed by $\boldsymbol{p}$ and $\boldsymbol{q}$. And we can model the above problem with the following bilevel polynomial optimization formulation.
\begin{equation}
\begin{array}
{ll}\min& \mathcal{F}(\boldsymbol{q},\boldsymbol{p}|\boldsymbol{X}_{int},\boldsymbol{X}_{obs})  \\
\text{s.t.}& \mathcal{f}_{i}(\boldsymbol{q},\boldsymbol{p})\leq0,\quad i=1,\ldots,I  \\
&\boldsymbol{p}=\operatorname{argmin}_{\boldsymbol{p}^{\prime}}\mathcal{G}(\boldsymbol{q},\boldsymbol{p^{\prime}}|X_{obs}) \\
&\mathrm{s.t.}\quad \mathcal{g}_{j}(\boldsymbol{p})\leq0,\quad j=1,\ldots,J \\
\mathrm{var~}&\boldsymbol{q},\boldsymbol{p}{,}
\end{array}
    \label{e11}
\end{equation}
where $\boldsymbol{X}_{obs}\in\mathbb{R}^{N_{obs}\times D}$ is the observational data with $D$ representing the number of random variables. $\boldsymbol{X}_{int}^{I_t}\in\mathbb{R}^{N_{int}\times D}$ represents the interventional data, where interventions are performed on variable $t\subset[D]$. And the interventional data under different variables is denoted as $\boldsymbol{X}_{int}=\{\boldsymbol{X}_{int}^{I_{t}}\}_{t=1}^{D}$. $\mathcal{F}$ and $\mathcal{f}_{i}$ denote the polynomial objective function and polynomial constraints of the upper-level problem, respectively. The function $\mathcal{F}$ is defined as the least squares loss, $\mathcal{F}=\mathcal{L}(\boldsymbol{q},\boldsymbol{p}|\boldsymbol{X}_{int},\boldsymbol{X}_{obs})=\sum_{t=1}^{D}\mathcal{L}(\boldsymbol{q},\boldsymbol{p}|\boldsymbol{X}_{int}^{I_{t}})+\alpha\mathcal{L}(\boldsymbol{q},\boldsymbol{p}|\boldsymbol{X}_{obs})$  ($\alpha$ is an empirical parameter, relevant details are given in the Appendix B), and the function $\mathcal{f}_{i}=\mathcal{h}_{i}(q,p)$ represents an acyclic constraint with different step lengths. Here, we use both observational and interventional data to learn variable $\boldsymbol{q}$ in the upper-level function, which not only fully leverages the information from observational data, but also improves the accuracy of results by incorporating interventional data. When calculating the upper-level objective function, we masked the intervened variable because the parent set of it changes during perfect intervention. $\mathcal{G}$ and $\mathcal{g}_{j}$ represent the polynomial objective function and polynomial constraints of the lower-level problem respectively, with $\mathcal{G}={\cal L}(\boldsymbol{q},\boldsymbol{p}|\boldsymbol{X}_{obs})+\lambda_{sp}{\cal L}_{sp}(\boldsymbol{p})$ and $\mathcal{g}_{j}(\boldsymbol{p})=-p_{j}^2(1-p_{j}^2)$. ${\cal L}_{sp}$ is the regularization function. Moreover, due to the sparsity of the causal structure, in most scenarios, we can simplify $\boldsymbol{q}$ and $\boldsymbol{p}$ by preprocessing or even introducing expert knowledge. For example, when there is obviously no causal relationship between two variables, the corresponding edge $p_i=0$.

The proposed method can effectively utilize observational and interventional data, and improve the accuracy of causal structure learning. For some scenarios where there is no or limited intervention data, we can utilize data augmentation techniques to simulate the interventions to improve the usability of our method. M. Ilse \cite{ilse2021selecting} introduced Intervention-augmentation equivalence (IAE), demonstrating the feasibility of simulating interventional data through data augmentation using only observational data under IAE conditions. In an IAE causal process $f_{X}{:}\,{\mathcal{D}}\times \mathcal{Y}\to{\mathcal{X}}$, every stochastic data augmentation transformation $\operatorname{aug}(\cdot)$ on $x\in \mathcal{X}$ is equivalent to a corresponding noise intervention $do(\cdot)$ on $d\in \mathcal{D}$ such that: $\operatorname{aug}(f_{X}(d,y))=f_{X}(do(d),y)$. Hence, by verifying if the causal relationship between variables meets the IAE condition, an appropriate data augmentation model can be trained for each variable and simulate interventions through the addition of noise. More details can be found in the original publication.

\textbf{Reformulation based on optimal conditions.} Although Eq. (\ref{e11}) is a bilevel polynomial optimization problem with both upper and lower constraints, fortunately, the lower optimization problem can be proved convex because of the operations that preserve convexity \cite{boyd2004convex}. According to \textit{Proposition 1}, since the lower-level problem satisfies the Slater's condition (\textit{Definition 1}), Eq. (\ref{e11}) can be transformed into a single-level polynomial optimization problem.
\begin{equation}
\begin{array}
{ll}\min &\mathcal{F}(\boldsymbol{q},\boldsymbol{p},\boldsymbol{\lambda}|\boldsymbol{X}_{int},\boldsymbol{X}_{obs})  \\
\text{s.t.} &\mathcal{f}_{i}(\boldsymbol{q},\boldsymbol{p})\leq0,\quad i=1,\ldots,I  \\
&\lambda_{0}\nabla_{\boldsymbol{p}}{\cal G}(\boldsymbol{q},\boldsymbol{p}|\boldsymbol{X}_{obs})+\sum_{j=1}^{J}\lambda_{j}\nabla_{\boldsymbol{p}}\mathcal{g}_{j}(\boldsymbol{p})=0 \\
&\lambda_{0},\lambda_{1},\ldots,\lambda_{J}\geq0 \\
&\mathcal{g}_{j}(\boldsymbol{p})\leq0,\lambda_{j}\mathcal{g}_{j}(\boldsymbol{p})=0,\quad j=1,\ldots,J \\
\mathrm{var~}&\boldsymbol{q},\boldsymbol{p},\boldsymbol{\lambda}{.} 
\end{array}
    \label{e12}
\end{equation}

We denote the inequality constraints and equality constraints as follows:

$\hat{\mathcal{f}}_{k_1}(\boldsymbol{q},\boldsymbol{p},\boldsymbol{\lambda})=\\ \quad\quad\begin{cases}-\mathcal{f}_{k_1}(\boldsymbol{q},\boldsymbol{p}),&k_1=1,...,I\\\lambda_{k_1-1},&k_1=I+1,...,I+J+1\\-\mathcal{g}_j(\boldsymbol{p}),&k_1=I+J+2,...,I+2J+1\end{cases}$ and 

$\widehat{\mathcal{g}}_{k_{2}}(\boldsymbol{q},\boldsymbol{p},\boldsymbol{\lambda})=\\ \quad\quad \begin{cases}\left(\lambda_{0}\nabla_{\boldsymbol{p}}\mathcal{G}(\boldsymbol{q},\boldsymbol{p}|\boldsymbol{X}_{obs})+\sum_{j=1}^{J}\lambda_{j}\nabla_{\boldsymbol{p}}\mathcal{g}_{j}(\boldsymbol{p})\right)_{k_{2}},\\\quad\quad\quad\quad\quad\quad\quad\quad\quad\quad\quad\quad\quad k_{2}=1,...,\overline{D}\\\lambda_{j}\mathcal{g}_{j}(\boldsymbol{p}),\quad\quad\quad\quad\quad\quad\quad\quad k_{2}=\overline{D}+1,...,\overline{D}+J{,}\end{cases}$

 then the problem Eq. (\ref{e12}) can be expressed as
\begin{equation}
\begin{array}
{ll}\min &\mathcal{F}(\boldsymbol{q},\boldsymbol{p},\boldsymbol{\lambda}|\boldsymbol{X}_{int},\boldsymbol{X}_{obs})  \\
\text{s.t.} &\hat{\mathcal{f}}_{k_{1}}(\boldsymbol{q},\boldsymbol{p},\boldsymbol{\lambda})\geq0,\quad  k_{1}=1,\ldots,I+2J+1  \\
&\widehat{\mathcal{g}}_{k_{2}}(\boldsymbol{q},\boldsymbol{p},\boldsymbol{\lambda})=0,\quad k_{2}=1,\ldots,\overline{D}+J \\
\mathrm{var~} &\boldsymbol{q},\boldsymbol{p},\boldsymbol{\lambda}{.}  
\end{array}
    \label{e13}
\end{equation}

\begin{proposition}[Equivalent single-level reformulation] 
Consider Eq. (\ref{e11}) where the lower-level optimization problem is convex and satisfies the slater condition. Then $\left(\boldsymbol{q}, \boldsymbol{p}\right)$ is a global optimal solution of the bilevel polynomial optimization problem Eq. (\ref{e11}) when there exist Lagrange multipliers $\boldsymbol{\lambda} \in \mathbb{R} ^{J+1}$ such that $\left(\boldsymbol{q}, \boldsymbol{p}, \boldsymbol{\lambda}\right)$ is the optimal solution to the single-level problem Eq. (\ref{e12})
\end{proposition}

\textit{Proof}. In order to prove the above proposition, we need to show that for any $\boldsymbol{q}$, there exists $\boldsymbol{\lambda} \in \mathbb{R} ^{j+1}$ such that the feasible set $\boldsymbol{p} \in \mathbb{P} \left(\boldsymbol{q} \right) $ of the lower optimization problem is equivalent to the following condition that

\begin{equation}
\resizebox{.85\linewidth}{!}{$
            \displaystyle
\begin{array}
{ll}
&\lambda_{0}\nabla_{\boldsymbol{p}}{\cal G}(\boldsymbol{q},\boldsymbol{p}|\boldsymbol{X}_{obs})+\sum_{j=1}^{J}\lambda_{j}\nabla_{\boldsymbol{p}}\mathcal{g}_{j}(\boldsymbol{p})=0 \\
&\lambda_{0},\lambda_{1},\ldots,\lambda_{J}\geq0 \\
&\mathcal{g}_{j}(\boldsymbol{p})\leq0,\lambda_{j}\mathcal{g}_{j}(\boldsymbol{p})=0,\quad j=1,\ldots,J {.} 
\end{array}$}
    \label{proof1-1}
\end{equation}

According to the \textit{Theorem 2.1} \cite{lasserre2010representations}, under the slater condition, a point $\boldsymbol{p}$ is the global optimal point of the lower-level problem if and only if $\boldsymbol{p}$ is the KKT point, i.e., there is $a_j \geq 0, j=1,...,J$ such that

\begin{equation}
\resizebox{.88\linewidth}{!}{$
            \displaystyle
\begin{array}
{ll}
&\nabla_{\boldsymbol{p}}{\cal G}(\boldsymbol{q},\boldsymbol{p}|\boldsymbol{X}_{obs})+\sum_{j=1}^{J}a_{j}\nabla_{\boldsymbol{p}}\mathcal{g}_{j}(\boldsymbol{p})=0 \\
&a_{j}\geq0, \mathcal{g}_{j}(\boldsymbol{p})\leq0, a_{j}\mathcal{g}_{j}(\boldsymbol{p})=0,\quad j=1,\ldots,J {.} 
\end{array}$}
    \label{proof1-2}
\end{equation}

Then we have $\lambda_0=\frac{1}{\sqrt{1+\sum_{j=1}^{J}a_j^2}}$ and $\lambda_j=\frac{a_j}{\sqrt{1+\sum_{j=1}^{J}a_j^2}}$ ($j=1,...,J$), where $a_j>0$. $\square$

\subsection{SDP Relaxations with Structured Sparsity}
In this subsection, we investigate how to solve the above single-level polynomial optimization problem Eq. (\ref{e13}), which can be outlined into three steps. First, the variables are partitioned according to the correlative sparsity; then, the term sparsity graph is constructed according to the correlation between the monomials; finally, the SDP relaxation problem of the POP problem is provided and solved.

\paragraph{Generate Cliques} Let $\widehat{\boldsymbol{x}}=(\boldsymbol{q},\boldsymbol{p},\boldsymbol{\lambda})$. First, we need to divide the variables involved in the POP problem into different cliques to reduce the scale of the final SDP problem. In Eq. (\ref{e12}), since $q_i$ and $p_i$ jointly define the causal structure, they have co-occurrence in the polynomial function $\mathcal{F}$ and the acyclic constraint function $\mathcal{f}_i$. In addition, $\lambda_i$ and $p_i$ also have co-occurrence in the constraint function $\lambda_{i}\mathcal{g}_{i}(\boldsymbol{p})=0$. Therefore, we can quickly divide the variables by the following steps:

\begin{itemize}
\item Construct the CSP graph for variable $\boldsymbol{p}$, denoted as $\mathbb{G}_{\boldsymbol{p}}^{\mathrm{csp}}(V,E)$, with nodes $V=[\bar{D}]$ corresponding to variable $\boldsymbol{p}$. For variables $p_{i_1}$ and $p_{i_2}$, the edge $(i_1,i_2 )\in E$ exists when one of the following two conditions is satisfied:

\begin{itemize}
    \item Let $\mathrm{supp}_{\boldsymbol{p}}(\mathcal{F})$ denote the support of the variable $\boldsymbol{p}$ in the polynomial function $\mathcal{F}$. For $\boldsymbol{\alpha} \in \mathrm{supp}_{\boldsymbol{p}}(\mathcal{F})$, there are $\alpha_{i_1}>0$ and $\alpha_{i_2}>0$;
    \item If both variables $p_{i_1}$ and $p_{i_2}$ are involved in the identical constraint function of Eq. (\ref{e13}).
\end{itemize}
\item The chordal extension of $\mathbb{G}_{\boldsymbol{p}}^{\mathrm{csp}}$ is denoted as $(\mathbb{G}_{\boldsymbol{p}}^{\mathrm{csp}})^{\prime}$ and the variables in $(\mathbb{G}_{\boldsymbol{p}}^{\mathrm{csp}})^{\prime}$ are partitioned and denoted as cliques, $\mathbb{C}=\{C_{l}\}_{l=1}^{L}$, with $\boldsymbol{p}[C_{l}]\subseteq \boldsymbol{p}$;
\item Based on the co-occurrence relationship between the variables, the cliques are extended to $\mathbb{C}'=\{C_{l}^{\prime}\}_{l=1}^{L}$, and ${\widehat{\boldsymbol{x}}}[C_{l}^{\prime}]\subseteq{\widehat{\boldsymbol{x}}}$.
\end{itemize}
    
\paragraph{Generate Term Sparsity} We first generate the standard monomial basis for each clique $C_l^{\prime}$ and obtain the term sparsity pattern by analyzing the correlations between monomials. Let $A:=\mathrm{supp}({\cal F})\cup\cup{\mathrm{supp}\widehat{\mathcal{f}}}_{k_{1}}\cup\cup\mathrm{supp}\widehat{\mathcal{g}}_{k_{z}}$  denote supports in the problem, and $A_l \left(l=1,\dots,L\right)\in A$ be the supports w.r.t $C_l^{\prime}$. For a relaxation order of $d\geq d_{\mathrm{min}}$ and $j\in\{0\}\cup J_{l}$, $\mathbb{G}_{d,l,j}^{\mathrm{tsp}}(V_{d,l},E_{d,l})$ denotes the TSP subgraph corresponding to the clique $C_l^{\prime}$, with the nodes $V_{d,l}$ corresponding to the monomial basis $\mathbb{N}_{d}^{n_{l}}$. The edges $E$ in $\mathbb{G}_{d,l,j}^{\mathrm{tsp}}$ is obtained through two successive steps in Section 2, including support extension and chordal extension. We denote the adjacency matrix of the TSP graphs as $\boldsymbol{B}_{d,l}^{\mathrm{tsp}}$, $\boldsymbol{B}_{d,l,j}^{\mathrm{tsp}}$.
   
\paragraph{SDP Relaxation Problem of POP} Let $\boldsymbol{y}$ denote the univariate vector corresponding to the standard monomial basis $\mathbb{N}_{d}^{N}$, and the linear function is defined as $\mathcal{L}_{\boldsymbol{y}}\colon\mathbb{R}[\widehat{\boldsymbol{x}}]\to\mathbb{R}$. The order of the functions is defined as $d_{k_{1}}=\lceil\operatorname{deg}(\hat{\mathcal{f}}_{k_{1}})/2\rceil$, $d_{k_{2}}=\lceil\operatorname{deg}(\hat{\mathcal{g}}_{k_{2}})/2\rceil$, $d_{\mathrm{min}}=\max\{\lceil\mathrm{deg}(\mathcal{F})/2\rceil,d_{k_{1}=1,\ldots,I+2J+1},d_{k_{2}=1,\ldots,\bar{D}+J}\}$. According to \cite{waki2006sums}, we set the relaxation order as $d\geq d_{\mathrm{min}}$.

We can obtain the moment submatrix $\boldsymbol{M}_{d}(\mathbf{y},C_{l}^{\prime})$ for each clique $C_{l}^{\prime}$ by the following equation:
    
\begin{equation}
    \boldsymbol{M}_{d}(\boldsymbol{y},C_{l}^{\prime})_{ij}=\mathcal{L}_{\boldsymbol{y}}\big(\mathbb{N}_{d}^{n_{l}}\left[i\right]\mathbb{N}_{d}^{n_{l}}\left[j\right]\big)=y_{i}y_{j}{,}
    \label{e14}
\end{equation}
where $i$, $j$ are the indexes of the matrix, and $\boldsymbol{M}_{d}(\boldsymbol{y},C_{l}^{\prime})_{0,0}$ is set as 1.

For $\hat{\mathcal{f}}_{k}(k\in J_{l})$, define $\boldsymbol{M}_{d-d_{k}}(\hat{\mathcal{f}}_{k}\boldsymbol{y},C_{l}^{\prime})$ as the localizing submatrix, where the entries in the matrix are derived from the following equation:
$$\boldsymbol{M}_{d-d_{k}}\big(\hat{\mathcal{f}}_{k}\boldsymbol{y},C_{l}^{\prime}\big)_{ij}=\mathcal{L}_{y}\big(\hat{\mathcal{f}}_{k}\mathbb{N}_{d-d_{k}}^{n_{l}}[i]\mathbb{N}_{d-d_{k}}^{n_{l}}[j]\big)\\$$
\begin{equation}
    \quad\quad\quad=\sum_{\alpha\in A_{l}}y_{\alpha}y_{i}y_{j}{.}
    \label{e15}
\end{equation}
    
The POP problem can then be relaxed to an SDP problem by introducing moment submatrices and localizing submatrices, shown as follows.
    
\begin{equation}
    \begin{aligned}
    &\text{min}&& \mathcal{L}_{\boldsymbol{y}}\big(\mathcal{F}(\widehat{\boldsymbol{x}}|\boldsymbol{X}_{int},\boldsymbol{X}_{obs})\big)  \\
    &\text{s.t}&& \boldsymbol{B}_{d,l}^{\mathrm{tsp}}\odot \boldsymbol{M}_{d}(\boldsymbol{y},C_{l}^{\prime})\in\prod_{\mathbb{G}_{d,l}^{\mathrm{tsp}}}(\mathbb{S}_{+}),l=1,\dots L  \\
    &&&\boldsymbol{B}_{d,l,j}^{\mathrm{tsp}}\odot\hat{\mathcal{f}}_{j}(\widehat{x})\in\prod_{\mathbb{G}_{d,l,j}^{\mathrm{tsp}}}(\mathbb{S}_{+}),\quad j\in J_{l},l=1,\ldots L \\
    &&&\boldsymbol{B}_{d,l,j}^{\mathrm{tsp}}\odot\widehat{\mathcal{g}}_{j}(\widehat{x})\in\prod_{\mathbb{G}_{d,l,j}^{\mathrm{tsp}}}(\mathbb{S}_{+}),\quad j\in J_{l},l=1,\dots L \\
    &\text{var}&& \widehat{\boldsymbol{x}}{.}
    \end{aligned}
    \label{e16}
\end{equation}

\begin{proposition}[Monotonic convergence]
For any relaxation order $d\geq d_{\mathrm{min}}$, the optimal value of the problem Eq. (\ref{e13}) monotonically converges to the optimal value of the original problem Eq. (\ref{e11}) as $d$ increases.
\end{proposition}

We give the proof of Proposition 2 in Appendix A.

It shows that by using different orders of semidefinite relaxation, the global optimal solution of the original problem can be gradually approximated. This provides sufficient theoretical guarantee for the global optimality of Bloom in solving the POP problems given in Eq. (\ref{e11}). In addition, there exist efficient interior point method \cite{bai2008semidefinite} for SDP problems. It has been shown that theoretically the interior point method can provide a global optimal solution with a quadratic convergence rate \cite{potra2000interior, zhang2020rate, yamashita2005quadratic}. In contrast, gradient descent algorithm often exhibit sublinear convergence rate for optimization problem with non-convex objective function \cite{li2023faster, zou2019sufficient, johnson2013accelerating}. Therefore, the proposed method is guaranteed to converge. In addition it exhibits faster convergence speed compared to gradient descent-type methods. 

\begin{proposition}[Optimality condition]
If solution to Eq. (\ref{e16}) satisfies the equivalence constraints as follows, 1) $\mathcal{f}_{i}(\boldsymbol{q},\boldsymbol{p})=0$ $(i=1,\ldots,I)$; 2) $\mathcal{g}_{j}(\boldsymbol{p})=0$ $(j=1,\ldots,J)$, then the obtained solution is the optimal solution to the original problem given in Eq. (\ref{e11}).
\end{proposition}

\textit{Proof}. We denote the optimal solution of the problem Eq. (\ref{e16}) under the relaxation order $d$ as $val_d$, since as the relaxation order increases, it gradually approaches the optimal solution of the original problem \cite{waki2006sums}, that is:
$$
    val_{d_{min}} \rightarrow val_{d_{min}+1} \rightarrow \dots \rightarrow val_{d} \dots \rightarrow val_{Eq. (\ref{e11})}{.}
$$
Theoretically, the optimal solution will asymptotically converge to the global optimal solution of the original problem. But the most original optimization problem is a discrete problem, with $\boldsymbol{q} \in \left\{0, 1\right\}^D$ and $\boldsymbol{p} \in \left\{-1, 0, 1\right\}^D$. Thus, $\mathcal{f}_{i}(\boldsymbol{q})=q_i\times(q_i-1)=0$ and $\mathcal{g}_{j}(\boldsymbol{p})=p_j^2\times(p_j^2-1)=0$.By converting equality constraints into inequality constraints, we relax its domain to $\left[-1, 1 \right]^D$ and $\left[0, 1 \right]^D$ so that it can be continuously optimized. Therefore, when the final solution $val_{d}$ $(d>d_{min})$ satisfies the equality constraints, i.e., the obtained $\boldsymbol{p}$ is exactly {-1, 0, 1} vector and  $\boldsymbol{q}$ is {0, 1} vector, the current solution can be considered optimal.$\square$

In addition, according to \cite{lasserrearticle}, under sufficient rank conditions, we can also determine whether the optimal value under the current relaxation order is the optimal value of the original problem by checking whether finite convergence has occurred. The steps of Bloom are shown in Algorithm \ref{alg1}.

\begin{algorithm}[htb]
\caption{Framework of Bloom.}
\label{alg1}
\begin{algorithmic}[1] 
\REQUIRE ~~\\ 
    The observational data, $\boldsymbol{X}_{obs}$;\\
    The interventional data, $\boldsymbol{X}_{int}$;\\
    Iteration, $t=0$; SDP relaxation order, $d$;\\
    Initial variables, $\boldsymbol{q}_0,\boldsymbol{p}_0,\boldsymbol{\lambda}_0$;
\ENSURE ~~\\ 
    Learned DAG, $\boldsymbol{W}$;
    \STATE Construct bilevel polynomial optimization formulation Eq. (\ref{e11}) with $\boldsymbol{q}_0$, $\boldsymbol{p}_0$, $\boldsymbol{X}_{obs}$ and $\boldsymbol{X}_{int}$;
    \STATE Calculate the optimal conditions of the lower-level problem and reformulate as Eq. (\ref{e13});
    \STATE Generate cliques and term sparsity of variables in Eq. (\ref{e11});
    \STATE Generate moment submatrices and localizing submatrices with Eq. (\ref{e14}) and Eq. (\ref{e15}), and reformulate as the SDP relaxation problem Eq. (\ref{e16});
    \FOR{$t=0$; $t<T$; $t++$}
    \STATE Solve Eq. (\ref{e16}) with Interior point method;
    \ENDFOR
    \RETURN $\boldsymbol{q}$, $\boldsymbol{p} \rightarrow \boldsymbol{W}$
\end{algorithmic}
\end{algorithm}

\subsection{Distributed Bloom}

However, due to privacy concerns, local datasets are not allowed to be uploaded to a central server in some scenarios \cite{mian22a, gao2021feddag, li2024federated}. Therefore, we attempt to further extend the Bloom algorithm to distributed settings, enabling it to learn the graph structure from distributed data without sharing locally stored data.

\textbf{Distributed Data}. Let $C=\{c_1,c_2,...,c_M\}$ be the client set which includes $M$ different clients, and $S$ represent the central server. The dataset $X^{c_m}=\{\boldsymbol{X}^{c_m}_{obs}, \boldsymbol{X}^{c_m}_{int}\}$ represent the local data owned by the client $c_m$. The dataset $X=\{X^{c_1},X^{c_2},...,X^{c_M}\}$ is called distributed dataset. And we define $X$ as a homogeneous distributed dateset, which means they are sampled from an identical distribution.

To learn causal structures from distributed data, the distributed Bloom solves each subproblem Eq. (\ref{e_dis}) by distributing it across all local clients. Since data is not shared between clients and the server, data privacy is significantly ensured. During training, the server and clients will exchange updated variables in each communication round to facilitate coordinated joint learning of the causal structure.

\begin{equation}
    \begin{aligned}
    &\text{min}&& \mathcal{L}_{\boldsymbol{y}}\big(\mathcal{F}(\widehat{\boldsymbol{x}}^{c_m}|X^{c_m})\big)  \\
    &\text{s.t}&& \boldsymbol{B}_{d,l}^{\mathrm{tsp}}\odot \boldsymbol{M}_{d}(\boldsymbol{y}^{c_m},C_{l}^{\prime})\in\prod_{\mathbb{G}_{d,l}^{\mathrm{tsp}}}(\mathbb{S}_{+}),l=1,\dots L  \\
    &&&\boldsymbol{B}_{d,l,j}^{\mathrm{tsp}}\odot\hat{\mathcal{f}}_{j}(\widehat{x}^{c_m})\in\prod_{\mathbb{G}_{d,l,j}^{\mathrm{tsp}}}(\mathbb{S}_{+}),\quad j\in J_{l},l=1,\ldots L \\
    &&&\boldsymbol{B}_{d,l,j}^{\mathrm{tsp}}\odot\widehat{\mathcal{g}}_{j}(\widehat{x}^{c_m})\in\prod_{\mathbb{G}_{d,l,j}^{\mathrm{tsp}}}(\mathbb{S}_{+}),\quad j\in J_{l},l=1,\dots L \\
    &\text{var}&& \widehat{\boldsymbol{x}}^{c_m}{.}
    \end{aligned}
    \label{e_dis}
\end{equation}

\textbf{Client Update.} Essentially, solving the optimization problem on each client can be seen as an independent process. Therefore, each client will calculate sub-problems with local dataset, following the steps in Algorithm 1. After each communication round, the client will receive the updated variables, which will be used as the initial values for the next round of solving problem.

\textbf{Server Update.} After clients completed local updates, the server randomly selects $r$ clients and collects the learned variables to the set $\widehat{x}_r=\{\widehat{\boldsymbol{x}}^{c_1}, \widehat{\boldsymbol{x}}^{c_2}, ..., \widehat{\boldsymbol{x}}^{c_r}\}$. By averaging the collected variables, we will get $\widehat{\boldsymbol{x}}_{new}$, which is then distributed to all clients.

The specific steps are shown in the Algorithm \ref{alg2}.

\textbf{Privacy Protection.} To avoid leakage of raw data in the client, distributed Bloom only exchanges the learned variables during the training process. Therefore, we consider the information leakage of local data to be relatively limited. For the graph structure information that may be contained in the transmission, we can address it by selecting a client as a proxy server \cite{gao2021feddag}. It is worth mentioning that our work only provides a possible distributed extension of Bloom. Regarding further privacy protection efforts, we could introduce more advanced privacy protection techniques \cite{9048613}, which will be the focus of future research.

\textbf{Communication Cost.} Distributed Bloom only requires exchanging parameters between the server and clients during communication rounds. Despite introducing some communication costs, we consider this to be acceptable with relatively low communication overhead. In each communication round, servers only need to collect learned parameters from selected clients and distribute updated variables to each client. Furthermore, the trade-off between performance and communication costs can also be controlled by choosing the number of selected clients $r$.

Additionally, for some large-scale structure learning problems, solving each subproblem still imposes significant demands on the computational resources of individual clients \cite{brouillard2020differentiable}. Fortunately, due to the sparsity of the constructed SDP problems Eq.(\ref{e16}), we can apply many existing distributed solving methods \cite{disSDP1, disSDP2, disSDP3}. 
Although this is not the main focus of this article, we also present a feasible approach in Appendix B.

\begin{algorithm}[htb]
\caption{Framework of Distributed Bloom.}
\label{alg2}
\begin{algorithmic}[1] 
\REQUIRE ~~\\ 
    The distributed dataset, $X=\{X^{c_1},X^{c_2},...,X^{c_M}\}$;\\
    Local iteration, $t^{c_m}=0$; Server iteration, $t^{s}=0$\\
    Communication round, $t_{cr}$\\
    SDP relaxation order, $d^{c_1}$; Selected clients, $r$;\\
    Initial variables, $\boldsymbol{q}_0^{c_1},\boldsymbol{p}_0^{c_1},\boldsymbol{\lambda}_0^{c_1}$;
\ENSURE ~~\\ 
    Learned DAG, $\boldsymbol{W}$;
    \FOR {$t^{s}=0$; $t^{s}<T^{s}$; $t^{s}++$}
        \FOR{\textbf{each} client $c_m$; $t^{c_m}=0$; $t^{c_m}<T^{c_m}$; $t^{c_m}++$}
        \STATE Solve Eq. (\ref{e_dis}) with Interior point method (similar to Algorithm \ref{alg1});
        \ENDFOR
        \IF{$t^{s} \% t_{cr}=0$ or $t^{s}=(T^{s}-1)$}
        \STATE Server collecting: randomly select $r$ client and collect their $\widehat{x}_r$;
        \STATE Server updating: aggregating and averaging $\widehat{x}_r$;
        \STATE Broadcasting the new variables $\widehat{\boldsymbol{x}}_{new}$;
        \FOR{\textbf{each} client $c_m$}
        \STATE Client updating: $\widehat{\boldsymbol{x}}_{c_m} \leftarrow \widehat{\boldsymbol{x}}_{new}$;
        \ENDFOR
        \ENDIF
    \ENDFOR
\RETURN $\widehat{\boldsymbol{x}}_{new} \rightarrow \boldsymbol{W}$; 
\end{algorithmic}
\end{algorithm}

\section{Experiments}

\subsection{Experimental Setup}
We conduct experiments on synthetic and real data respectively to demonstrate its effectiveness by comparing the proposed method with some typical methods. All experiments are implemented on a server with Intel(R) Xeon(R) CPU E5-2690 v3 CPUs.

\subsubsection{Datasets}
We first randomly generated multiple datasets based on different structural causal models. These models vary in the size of the graph size, edge sparsity, and causal relationships. We use a scale-free (SF) graphical model to draw a random DAG $\mathbb{G}$ and then generate data according to the causal order of $\mathbb{G}$. We mainly consider perfect interventions, where the conditional distribution of targeted nodes were replaced by a new distribution similarly to \cite{hauser2012characterization, squires2020permutation, brouillard2020differentiable}. In experiments, we assume that the observational data is sampled from the Gaussian distributions, while the intervented variables obey the uniform distributions. Since some baseline methods assume linear relationships, we conducted two experiments in which the relationship between nodes was linear or polynomial, respectively. The linear datasets are generated following $X_i=w_i^TX+0.4\cdot N_i$ ($i\in [D]$), where the coefficients $w_i$ are sampled uniformly from $\left[-1.0, -0.4\right]\cup\left[0.4, 1.0\right]$.
The nonlinear dataset are generated following a polynomial function $X_i=\mathcal{F}({\mathrm{Pa}(X_i)})+0.4\cdot N_i$, i.e., $X_i = 0.8\cdot (X_j)^3 + N_i$, where the degrees of the polynomial terms are randomly sampled from $\{1,2,3\}$, similar to \cite{brouillard2020differentiable}. The sample length of each observational and interventional datasets are $N=300$. Under the linear experimental setting, the number of nodes in the graph are $D=(5, 7, 9, 11, 13, \, and\ , 15)$, and the number of edges is $(1.3\sim2)\cdot D$. And in the polynomial experiments, the number of nodes are $D=(5, 10, 15)$. In addition, we further tested the experimental results of the proposed algorithm under imperfect interventions in Section B.3. Finally, we also evaluated the scalability of our method on larger scales of nodes. In the scalability experiments, the number of nodes was set to $D=(50, 75, and\ ,100)$, with edges proportional to $1\cdot D$. In all settings, we randomly sampled 5 graphs to compare the average performance of each algorithm.

We also tested the performance of different methods on a flow cytometry dataset from Sachs et al.\cite{sachs2005causal}. This dataset measures the expression levels of phosphorylated proteins and phospholipids in human cells. Interventions were conducted by using reagents to activate or inhibit the measured proteins. In the experiment, we utilize a subset of this dataset, focusing on data where the measured proteins were directly perturbed. This subset comprises 5846 measurements, with 1755 being observational data, and the rest corresponding to measurements under interventions on five different single-node targets (target proteins: Akt, PKC, PIP2, Mek, PIP3). We employed Sachs' consensus graph as the ground truth, consisting of 11 nodes and 17 edges. It is important to note that in this real-world dataset, the assumption of causal sufficiency may not hold, and the interventions may be considered imperfect.

\subsubsection{Baselines}
We compared our method with several methods based on observational or interventional data, including,

\begin{itemize}
    \item \textbf{LINGAM}: LINGAM (Linear Non-Gaussian Acyclic Model) is a method based on Functional Causal Model (FCM), which reveals causal relationships by testing whether there are linear and non-Gaussian functional relationships between variables.
    \item \textbf{PC}: PC is an efficient constraint-based method that employs conditional independence tests to capture potential causal dependencies among variables. We employ the Fisher-z test (p-value = 0.05) for examination in linear experiments, and Kernel-based Conditional Independence (KCI) test  (p-value = 0.05) \cite{zhang2011kernel}  in polynomial experiments.
    \item \textbf{GES}: GES is a score-based method, widely employed in various applications. Adopting a greedy approach, GES iteratively optimizes the estimated causal graph by computing the score function and adjusting potential edges. Here, we use Bayesian Information Criterion (BIC) score as the evaluation function in linear experiments, and apply generalized score function \cite{huang2018generalized} for polynomial experiments.
    \item \textbf{Notears}: Notears is a score-based method that characterizes acyclicity using an equality constraint, enabling its solution through continuous optimization techniques.
    \item \textbf{Sortnregress}: Sortnregress \cite{reisach2021beware} is a causal structure learning method for observational data. It utilizes ranking mechanisms and regression to effectively identify causal relationships between variables.
    \item \textbf{IGSP}: IGSP is a hybrid method that optimizes a score based on conditional independence tests. We apply KCI test in nonlinear experiments.
    \item \textbf{GIES}: GIES is a variant of GES designed for discovering causal relationships in observational and interventional data. GIES assumes the targets of the interventional data are known.
    \item \textbf{DCDI}: DCDI is a method based on continuous optimization that uses observational and interventional data and can be used to discover nonlinear causal relationships.
    \item \textbf{ENCO}: ENCO is also a nonlinear causal structure learning method with observational and interventional data.
\end{itemize}

Among the aforementioned methods, LINGAM, PC, GES, Notears and Sortnregress exclusively rely on observational data, whereas IGSP, GIES DCDI and ENCO are adept at handling both observational and intervention data. In experiments on synthetic data, we evaluate the best performance of each method across different experimental settings. But in real-world data, we only compared the best performance of those methods using both observational and intervention data.

\subsubsection{Evaluation Metrics}We report True positive rate (TPR) and SHD (Structural Hamming Distance) to evaluate the quality of structure learning. SHD is the minimum number of edge additions, deletions, and inversions required to convert an estimated graph into a true DAG. It takes into account both false positives and false negatives, and a lower SHD indicates a better estimate.

\subsection{Experimental Results}
We tested the overall performance of all methods on two synthetic datasets and real datasets respectively.

\subsubsection{Performance Comparison on Synthetic Dataset}

We first conducted experiments under linear settings, and all results are displayed in Table \ref{tab:table2}. Since some of the baseline methods could not employ interventional data, for the purpose of comparison, we experimented the proposed Bloom on observational data only by using a single-level polynomial optimization, and named it Bloom(obs). The experimental results consistently demonstrate Bloom's superior performance over all baseline methods, highlighting the effectiveness of the proposed approach. And our method performs well when using only observational data, even outperforming methods that utilize interventional data in certain scenarios. Furthermore, Bloom excels in TPR and SHD across all scenarios. This indicates the superior ability of our method to estimate causal graphs, benefiting from the bilevel optimization framework that effectively captures causal information in various datasets. By formulating with POP, it allows our method to converge to an approximate global optimum through global search methods, ensuring a more accurate recovery of causal graphs. In contrast, GIES employs discrete greedy search methods, while most continuous optimization-based methods utilize gradient descent, which may be trapped in local optima. Additionally, the introduction of interventional data enhances the performance by supplying additional information about direct causal relationships between variables, enabling better learning of DAGs. 

Table \ref{tab:tablePoly} shows the performance of the proposed method on polynomial datasets. In the experiments, our method demonstrated superior causal structure recovery capabilities. Because Bloom can effectively incorporate interventional data through bilevel optimization, which helps clearly identifying causal directions and reducing the identification of spurious relationships. Additionally, it utilizes polynomial optimization, which significantly avoids the problem of local optima and improves convergence, enabling our method to find better solutions in complex causal graph structures.  On the other hand, although GIES showed suboptimal performance overall, its greedy strategy can sometimes cause it to get stuck in local optima, affecting the final outcome. Surprisingly, methods based on neural networks did not perform well. The main reason may be their high data requirement, which makes it difficult to fully optimize the model under limited conditions. In real-world scenarios, it is often challenging to obtain sufficient data.

\begin{table*}[htb]
\centering
\caption{Results of the Linear Experiments \label{tab:table2}}
\begin{threeparttable}
\resizebox{\linewidth}{!}{
\renewcommand\arraystretch{2.0}
\begin{tabular}{cccccccccccc}
\toprule[0.4mm]
\specialrule{0em}{1.8pt}{1.8pt}
\midrule
\multicolumn{2}{c}{} &
  \textbf{GES} &
  \textbf{LINGAM} &
  \textbf{PC} &
  \textbf{NOTEARS} &
  \textbf{Sortnregress} &
  \textbf{IGSP*} &
  \textbf{GIES*} &
  \textbf{Bloom(obs)} &
  \textbf{Bloom*} \\ \hline
\multirow{6}{*}{\textbf{TPR}} &
  \textbf{5 Nodes} &
  0.5000±0.1250 &  0.4500±0.2739 &  0.6250±0.0884 &  0.6750±0.2271 &  0.8000±0.0685 &  0.3250±0.0612 &  0.8250±0.2437 &  {\underline{ 0.8750±0.1250}} &  \textbf{0.9500±0.0685} \\ 
  \cline{2-11} 
  & \textbf{7 Nodes} &  
  0.4400±0.1517 &  0.3800±0.0834 &  0.4000±0.0353 &
  0.7600±0.1140 &  0.8600±0.0894 &  0.4200±0.1720 &  0.8600±0.2191 &  {\underline{ 0.9000±0.1000}}  &  \textbf{0.9000±0.0000} \\
  \cline{2-11}  
  & \textbf{9 Nodes} &
  0.4833±0.1086 &  0.3167±0.1990 &  0.4667±0.1264 &  0.8167±0.2312 &  0.8667±0.0457 &  0.3833±0.1795 &  0.9000±0.0697 &  {\underline{ 0.9167±0.0589}}  &  \textbf{0.9334±0.0373} \\
  \cline{2-11}  
  & \textbf{11 Nodes} &
  0.5231±0.1480 &  0.2462±0.1141 &  0.6154±0.0769 &  0.8462±0.1216 &  0.8462±0.0769 &  0.4154±0.1427 &  0.8769±0.1771 &  {\underline{0.8847±0.0444}}  &  \textbf{0.8923±0.0421} \\
  \cline{2-11}  
  & \textbf{13 Nodes} &
  0.6133±0.1909 &  0.4133±0.1283 &  0.4533±0.1095 &  0.8933±0.1299 &  0.9333±0.0667 &  0.4933±0.0904 &  {\underline{ 0.9467±0.1193}} &  0.9333±0.0667  &  \textbf{0.9600±0.0365} \\
  \cline{2-11} 
  & \textbf{15 Nodes} & 
  0.4340±0.1839 & 0.4823±0.1044 & 0.5529±0.0670 & 0.8235±0.0832 & 0.8941±0.0873 & 0.6000±0.1596 & {\underline{ 0.9176±0.1289}} & 0.9028±0.0688 & \textbf{0.9412±0.0588}     \\ 
  \hline
  \multirow{6}{*}{\textbf{SHD}} & 
  \textbf{5 Nodes} & 
  4.6±1.1       & 4.8±3.0       & 3.0±0.7       & 2.6±1.8  & 2.2±0.4 & 5.8±0.7  & 1.8±2.5 & {\underline{ 1.6±1.1}}          & \textbf{0.4±0.5} \\ 
  \cline{2-11} 
  & \textbf{7 Nodes} & 
  8.4±2.3       & 7.8±0.8      & 6.0±0.0       & 2.4±1.1 & 2.8±0.8 & 2.8±1.7 & 2.0±3.5 & {\underline{ 1.6±1.9}}  & \textbf{1.0±0.0} \\ 
  \cline{2-11} 
  & \textbf{9 Nodes} & 
  8.8±1.9       & 10.2±3.0      & 6.4±1.5       & 2.2±2.8 & 3.0±1.9 & 8.0±2.4 & {\underline{ 1.8±1.1}} & 1.8±1.3             & \textbf{1.6±1.1}     \\ 
  \cline{2-11} 
  & \textbf{11 Nodes} & 
  11.0±4.4       & 12.6±0.5      & 5.0±1.0       & 2.2±1.9 & 4.2±1.9 & 8.6±2.2 & 3.0±4.1 & {\underline{ 2.4±0.5}}             & \textbf{1.6±0.5}     \\ 
  \cline{2-11} 
  & \textbf{13 Nodes} & 
  11.6±3.8       & 10.8±2.2      & 8.2±1.6       & {\underline{ 1.8±1.9}} & 2.4±0.9 & 9.6±1.6 & 1.6±3.6 & 2.4±1.8             & \textbf{2.0±1.4}     \\ 
  \cline{2-11} 
  & \textbf{15 Nodes} & 
  19.0±3.9       & 11.8±2.2      & 7.6±1.1       & 3.0±1.4 & 7.8±4.4 & 9.0±5.2 & {\underline{ 2.4±4.3}} & 2.8±0.8             & \textbf{1.0±1.0}     \\ 
\midrule
\specialrule{0em}{1.5pt}{1.5pt}
\toprule[0.4mm]
\end{tabular}}
\begin{tablenotes}
        \item \small $\bullet$ The marker * indicates that the method uses both observational and interventional data. 
        \item \small $\bullet$ The bold font in the table shows the best performance, while the underline font indicates the second best performance.
\end{tablenotes}
\end{threeparttable}
\end{table*}

\begin{table*}[htb]
\centering
\caption{Results of the Polynomial Experiments \label{tab:tablePoly}}
\begin{threeparttable}
\resizebox{0.85\linewidth}{!}{
\renewcommand\arraystretch{2.0}
\begin{tabular}{cccccccccc}
\toprule[0.4mm]
\specialrule{0em}{1.8pt}{1.8pt}
\midrule
\multicolumn{2}{c}{} &
  \textbf{PC} &
  \textbf{GES} &
  \textbf{GIES*} &
  \textbf{IGSP*} &
  \textbf{DCDI*} &
  \textbf{ENCO*} &
  \textbf{Bloom*} \\ \hline
\multirow{3}{*}{\textbf{TPR}} &
  \textbf{5 Nodes} &
  0.4000±0.4214 &  0.6286±0.1278 &  {\underline{ 0.6857±0.1195}} &  0.0.4000±0.2555 &  0.4857±0.4802 & 0.5000±0.3595 &  \textbf{0.7714±0.0782} \\ 
  \cline{2-9} 
  & \textbf{10 Nodes} &  
  0.4533±0.1282 &  0.5200±0.1095 &  {\underline{ 0.6133±0.1592}}  &  0.4267±0.1115 &  0.3333±0.0667 &  0.3867±0.1726 & \textbf{0.7200±0.0558} \\
  \cline{2-9}  
  & \textbf{15 Nodes} &
  0.4235±0.1341 &  0.3882±0.0789 &  {\underline{ 0.5294±0.0721}} &  0.3999±0.1206 &  0.2510±0.0479 &  0.3294±0.1354 &  \textbf{0.6823±0.1590} \\ 
  \hline
  \multirow{3}{*}{\textbf{SHD}} & 
  \textbf{5 Nodes} &
  4.8±2.8 &  3.0±1.2 &  {\underline{ 2.6±1.1}} &  4.6±2.3 &  5.0±1.9 &  4.8±1.5 &  \textbf{2.0±0.7}\\ 
  \cline{2-9} 
  & \textbf{10 Nodes} &
  14.4±4.0 &  {\underline{ 10.4±2.7}} &  12.6±5.6 &  10.6±2.6 &  14.8±1.3 &  12.4±1.5 &  \textbf{6.8±1.3} \\ 
  \cline{2-9} 
  & \textbf{15 Nodes} &
  18.6±5.7 &  22.6±1.1 &  18.6±2.6 &  {\underline{ 16.6±3.6}} &  19.6±2.1 &  17.6±4.0 &  \textbf{10.8±4.7}\\ 
\midrule
\specialrule{0em}{1.5pt}{1.5pt}
\toprule[0.4mm]
\end{tabular}}
\begin{tablenotes}
        \item \small $\bullet$ The marker * indicates that the method uses both observational and interventional data. 
        \item \small $\bullet$ The bold font in the table shows the best performance, while the underline font indicates the second best performance.
\end{tablenotes}
\end{threeparttable}
\end{table*}

To validate the effectiveness of the distributed Bloom in large-scale causal structure learning problems, we conducted experiments under 50, 75 and 100 nodes, respectively. When the number of variables increases, learning causal structures becomes more complex and difficult. In this experiment, we only compared all methods that use both observational data and intervention data, and the experimental results are presented in Table \ref{tab:table3}. The results clearly indicate that our method maintains relatively good performance in large-scale problems and significantly outperforms the baseline methods. Additionally, Bloom can effectively improve the accuracy and reliability of problem solving by using polynomial optimization modeling, and make the obtained solution close to the global optimal. We further demonstrate the scalability of the proposed method on larger scale nodes in the Appendix E.

\begin{table}[htb]
\centering
\caption{Results of the Scalability Experiments} \label{tab:table3}
\begin{threeparttable}
\resizebox{0.95\linewidth}{!}{
\renewcommand\arraystretch{1.65}
\begin{tabular}{cccccc}
\toprule[0.4mm]
\specialrule{0em}{1.5pt}{1.5pt}
\midrule
\multicolumn{2}{c}{} &
  \textbf{IGSP} &
  \textbf{GIES} &
  \textbf{Bloom} \\ \hline
\multirow{3}{*}{\textbf{TPR}} &
  50 Nodes &
  0.7000±0.1131 &  0.8520±0.0559 &  \textbf{0.9560±0.0167} \\   \cline{2-5} &
  75 Nodes &  
  0.7042±0.0740 &  0.9093±0.0520 & \textbf{0.9173±0.0318} \\   \cline{2-5} & 
  100 Nodes &
  0.7520±0.0286 &  0.8920±0.0228 & \textbf{0.9080±0.0465} \\   
  \hline
  \multirow{3}{*}{\textbf{SHD}} & 
  50 Nodes &
  17.4±5.6 &  18.0±8.9 &  \textbf{2.8±1.1} \\   \cline{2-5} & 
  75 Nodes &  
  36.4±11.3 &  28.2±10.4 & \textbf{7.6±3.6} \\   \cline{2-5}  & 
  100 Nodes &
  46.0±10.6 &  40.6±7.3 & \textbf{15.2±2.8} \\ 
\midrule
\specialrule{0em}{1.5pt}{1.5pt}
\toprule[0.4mm]
\end{tabular}}
\end{threeparttable}
\end{table}

\subsubsection{Performance Comparison on Real Dataset}

We tested our approach on the flow cytometry data set from Sachs et al., a commonly used dataset for causal structure learning problems. In Table \ref{tab:table4}, we report the SHD and TPR for all methods. Bloom performed exceptionally well, exhibiting the best overall performance with the lowest SHD and highest TPR. In contrast, other baseline methods often identify a considerable number of false or erroneous causal relationships. This is attributed to the increased complexity of causal relationships between variables in real-world scenarios, which might even be nonlinear, posing significant challenges for causal structure learning. However, experimental results demonstrate that our method exhibits robust causal inference performance in this real-world scenarios, and is capable of recovering causal structures more comprehensively.

\begin{table}[htb]
\centering
\footnotesize
\caption{Experimental Results on Real data \label{tab:table4}}
\begin{threeparttable}
\resizebox{0.48\linewidth}{!}{
\renewcommand\arraystretch{1.6}
\begin{tabular}{cccc}
\toprule[0.4mm]
\specialrule{0em}{1.5pt}{1.5pt}
\midrule
\multicolumn{1}{l}{} & \textbf{SHD} & \textbf{TPR}   \\ \hline
\textbf{IGSP}        & 18           & 0.4     \\ \hline
\textbf{GIES}        & 13           & 0.2941   \\ \hline
\textbf{DCDI}       & 33           & 0.3529          \\ \hline
\textbf{ENCO}       & 25           & 0.4118           \\ \hline
\textbf{Bloom}       & \textbf{5}   & \textbf{0.7059} \\ \hline
\specialrule{0em}{1.5pt}{1.5pt}
\toprule[0.4mm]
\end{tabular}}
    \end{threeparttable}
\end{table}

\subsubsection{Experiment with Imperfect Interventions}
\begin{figure*}[htb]
\centering
\includegraphics[width=1.6\columnwidth]{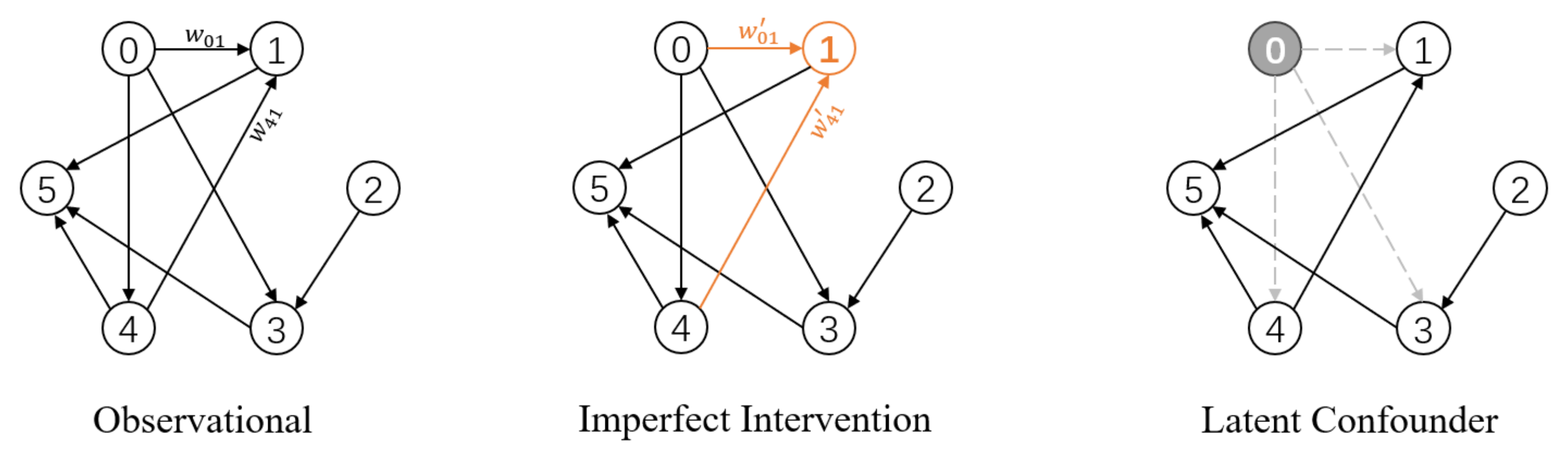}
\caption{Types of graphs under different experimental settings. In imperfect interventions, the weighted coefficients are changed. In latent confounder, a common cause is not observed.}
\label{fig_2}
\end{figure*}

In contrast to the previous experiments, we considered imperfect interventions in this study. As illustrated in the Fig. \ref{fig_2}, the causal relationships between the intervened variable and its parents is not entirely removed; rather, we adjust the causal strength between them. Here we perform a simple verification experiment. Specifically, we sample observational data from the following linear SEM: $X_i=X\boldsymbol{w}_i+z_i$, where the coefficients $W$ between variables are randomly sampled from the interval $\left[-1.0, -0.4\right]\cup\left[0.4, 1.0\right]$. Following \cite{brouillard2020differentiable}, for imperfect interventions, we modify the initial weighted coefficients of the intervened variable by sampling from a new value range of $\left[-2.0, -1.2\right)\cup\left(1.2, 2.0\right]$. For nodes without parents, the distributions of intervened nodes are replaced by uniform distributions. We randomly generated a set of five different graphs, each containing six observed variables and ten directed edges. The experimental results are shown in the Table \ref{tab:table5}. It is evident that even in the case of imperfect interventions, our method remains relatively effective in recovering causal structures.

\begin{table}[htb]
\centering
\footnotesize
\caption{Experimental Results on Imperfect Interventions \label{tab:table5}}
\begin{threeparttable}
\resizebox{0.54\linewidth}{!}{
\renewcommand\arraystretch{1.6}
\begin{tabular}{cccc}
\toprule[0.4mm]
\specialrule{0em}{1.5pt}{1.5pt}
\midrule
\multicolumn{1}{l}{} & \textbf{SHD} & \textbf{TPR}     \\ \hline
\textbf{IGSP}        & 5.8±3.1           & 0.5600±0.2245               \\ \hline
\textbf{GIES}        & 6.0±3.3           & 0.5800±0.2135               \\ \hline
\textbf{DCDI}       & 6.4±0.9           & 0.4200±0.0837               \\ \hline
\textbf{ENCO}        & 6.2±0.8           & 0.4200±0.1304               \\ \hline
\textbf{Bloom}       & \textbf{2.4±0.5}  & \textbf{0.7680±0.0271} \\ \hline
\specialrule{0em}{1.5pt}{1.5pt}
\toprule[0.4mm]
\end{tabular}}
    \end{threeparttable}
\end{table}

\subsubsection{Experiment with Latent Confounder}
In the previous section, we assumed sufficiency of the causal structure learning, implying that all variables in the causal relationship can be observed. However, in real-world scenarios, the presence of latent confounders is often inevitable. As shown in Fig. \ref{fig_2}, latent confounders are often unobserved common causal variables, which may introduce dependencies between two or more variables and lead to causal discovery methods identifying spurious causal relationships. Therefore, we intentionally relaxed this assumption and evaluated the performance of the proposed method in this subsection. 

We randomly generated a set of five graphs, each containing one latent confounder and five observed variables. In the experiment, we assumed that this unobserved variable is not a descendant of any other observed variables, and the data generation process remained consistent with the previous. Our objective was to accurately learn the causal structure among the observed variables. The experimental results, as shown in the Table \ref{tab:table6}, indicate that our method is effective in learning causal relationships between other variables even in the presence of latent confounders.  The inference performance of Bloom is closely associated with the introduction of intervention data and the modeling approach based on polynomial optimization.

\begin{table}[htb]
\centering
\footnotesize
\caption{Experimental Results on Latent Confounder \label{tab:table6}}
\begin{threeparttable}
\resizebox{0.54\linewidth}{!}{
\renewcommand\arraystretch{1.6}
\begin{tabular}{cccc}
\toprule[0.4mm]
\specialrule{0em}{1.5pt}{1.5pt}
\midrule
\multicolumn{1}{l}{} & \textbf{SHD} & \textbf{TPR}   \\ \hline
\textbf{IGSP}        & 3.6±1.4           & 0.4400±0.1497               \\ \hline
\textbf{GIES}        & 2.0±0.0           & 0.9600±0.0800               \\ \hline
\textbf{DCDI}       & 5.2±1.1           & 0.4000±0.1049               \\ \hline
\textbf{ENCO}       & 4.6±0.9           & 0.5000±0.1250  \\ \hline
\textbf{Bloom}       & \textbf{0.4±0.5}   & \textbf{1.0000±0.0000} \\ \hline
\specialrule{0em}{1.5pt}{1.5pt}
\toprule[0.4mm]
\end{tabular}}
    \end{threeparttable}
\end{table}

\subsubsection{Ablation Experiment}

To validate the benefits of incorporating interventional data into the modeling process, we further conducted experiments solely with observational data to assess the performance of model. The experiments were conducted on two distinct datasets comprising five and seven nodes, respectively. As represented in Table \ref{tab:table7}, the outcomes reveal that the inclusion of interventional data significantly enhances the model's capability to discover causalities and eliminate spurious relationships, thereby facilitating a more precise learning of the causal structure.

\begin{table}[htb]
\centering
\footnotesize
\caption{Results of Ablation Experimente \label{tab:table7}}
\begin{threeparttable}
\resizebox{\linewidth}{!}{
\renewcommand\arraystretch{2.0}
\begin{tabular}{ccccc}
\toprule[0.4mm]
\specialrule{0em}{1.5pt}{1.5pt}
\midrule
\multirow{2}{*}{}                & \multicolumn{2}{c}{\textbf{SHD}}    & \multicolumn{2}{c}{\textbf{TPR}}    \\ \cline{2-5} 
                                 & \textbf{5 Nodes} & \textbf{7 nodes} & \textbf{5 Nodes} & \textbf{7 nodes} \\ \hline
\textbf{\begin{tabular}[c]{@{}c@{}}Observational\\  Data Only\end{tabular}} & 0.8±0.4          & 2.2±0.8          & 0.8857±0.0639          & 0.8167±0.0697          \\ \hline
\textbf{Bloom}                   & \textbf{0.0±0.0} & \textbf{1.2±0.4} & \textbf{1.0000±0.0000} & \textbf{0.9000±0.0373} \\ \hline
\specialrule{0em}{1.5pt}{1.5pt}
\toprule[0.4mm]
\end{tabular}}
    \end{threeparttable}
\end{table}

\section{Conclusion}
This paper proposes Bloom for addressing the causal structure learning problem, which is based on bilevel polynomial optimization. This method efficiently integrates both observational and interventional data and can establish causal relationships with a higher level of confidence. Furthermore, we have extended it to a distributed setting for parallel processing across multiple distributed nodes. To our best knowledge, this study represents the inaugural exploration into employing bilevel polynomial optimization for causal structure discovery. Given the extensive literature on efficient algorithms as well as theoretical analyses of bilevel and polynomial optimization methods, we believe this novel perspective will pave new pathways for exploring the causal structure learning problem and contribute significantly to further research in this field.

\bibliographystyle{IEEEtran}
\bibliography{ref}

\clearpage

\appendix

\section{Proof}
\subsection{The Proof of Proposition 2}

\textbf{Proposition 2 [Monotonic convergence]:} For any relaxation order $d\geq d_{\mathrm{min}}$, the optimal value of the problem Eq.(20) monotonically converges to the optimal value of the original problem Eq.(11) as $d$ increases.

Before providing the proof of Proposition 2, we first give the following Lemma.

\begin{lemma}[Putinar’s Positivstellensatz \cite{Putinar}]
    Let $f_0$ and $f_i, i=1,...,p$ be real polynomials of $\boldsymbol{x} \in \mathbb{R}^v$. Suppose that there exist $R>0$ and sum-of-squares polynomials $\hat{\sigma}_1,...,\hat{\sigma}_p\in \sum^2 \left[\boldsymbol{x}\right]$ such that $R-||\boldsymbol{x}||^2=\hat{\sigma}_0(\boldsymbol{x}) + \sum_{i=1}^p \hat{\sigma}_i f_i(\boldsymbol{x})$ for all $\boldsymbol{x} \in \mathbb{R}^v$. If $f_0(\boldsymbol{x})>0$ over the set $\{\boldsymbol{x}\in \mathbb{R}^v:f_i(\boldsymbol{x})\geq0,i=1,...,p\}$, then there exist $\sigma_i\in \sum^2\left[\boldsymbol{x}\right], i=0,1,...,p$ such that $f_0=\sigma_0+\sum_{i=1}^p\sigma_i f_i$.
\end{lemma}

According to \cite{lasserre2001global}, we have the sum-of-square relaxation problems for Eq. (13) when relaxation order is $d$.

\begin{equation}
    \begin{aligned}
    \max \quad & \mu \\
    \text { s.t. } \quad & \mathcal{F}-\mu=\sigma_{0}-\sum_{k_1=1}^{I+2 J+1} \sigma_{k_1} \widehat{\mathcal{f}}_{k_1}-\sum_{k_2=1}^{\overline{D} + J} \phi_{k_2} \widehat{\mathcal{g}}_{k_2} \\
    & \sigma_{k_1} \in \Sigma^{2}[\boldsymbol{q}, \boldsymbol{p}, \boldsymbol{\lambda}],   \operatorname{deg} (\sigma_{0}) \leq 2 d, \\
    &\operatorname{deg}\left(\sigma_{k_1} \widehat{\mathcal{f}}_{k_1}\right) \leq 2 d, k_1=1, \ldots, I+2 J+1 \\
    & \phi_{k_2} \in \Sigma^2 [\boldsymbol{q}, \boldsymbol{p}, \boldsymbol{\lambda}], \operatorname{deg}\left(\phi_{k_2} \widehat{\mathcal{g}}_{k_2}\right) \leq 2 d, k_2=1, \ldots, \overline{D} + J\\
    \text {var.} \quad & \mu, \sigma_{k_1}, \phi_{k_2} {.}
    \end{aligned}
    \label{sos}
\end{equation}

And it is known the above question that can be reformulated as SDP problem \cite{lasserrearticle}, i.e., Eq. (18) in our problem.

\textit{Proof}. Let $val_d(Eq.(20))$ represent the optimal value of Eq.(20) with the relaxation order of $d$. According to Lasserre hierarchical theorem \cite{lasserre2001global}, it can be easily verified that $val_d(Eq.(20)) \leq val_{d+1}(Eq.(20)) \leq val_{Eq. (13)}$. Let $\epsilon >0$ and $d\in \mathbb{N}$, we define $\hat{\mathcal{F}}=\mathcal{F}-(val_{Eq.(13})-\epsilon)$. And $\hat{\mathcal{F}}>0$ over the feasible set. By applying Lemma 1, there exist $\sigma_{k_1}, \phi_{k_2}\in \sum^2[\boldsymbol{q}, \boldsymbol{p}, \boldsymbol{\lambda}]$ such that

$$\hat{\mathcal{F}}=\sigma_0-\sum_{k_1=1}^{I+2 J+1} \sigma_{k_1} \widehat{\mathcal{f}}_{k_1}-\sum_{k_2=1}^{\overline{D} + J} \phi_{k_2} \widehat{\mathcal{g}}_{k_2}{.}$$

Then, we have $val_{d}(Eq.(20))\geq val_{Eq.(13)}-\epsilon$. Since $val_{d}(Eq.(20)) \leq val_{Eq. (13)}$ for all $d \geq d_{\text{min}}$. Therefore, $val_{Eq.(18)}=val_{d}(Eq.(20))\rightarrow val_{Eq.(13)}=val_{Eq.(11)}$. $\square$

\section{Methodology}

\subsection{Distributed Bloom for Large-Scale Structure Learning Problem}

For causal structure learning problems, when the number of nodes is large, it will place high demands on the computing resources of a single client. Therefore, for each sub-problem on the client, we can further decouple it into multiple SDP sub-problems according to the sparsity of the SDP problem and solve them across distributed nodes.

Due to the sparsity of the problem, Eq. (19) can exhibit the following coupling form for sub-problem of each client (omit client superscript):
\begin{equation}
    \resizebox{.88\linewidth}{!}{$
            \displaystyle
    \begin{aligned}
    &\text{min}&&\sum_{l=1}^{L}\mathcal{L}_{\mathbf{y}}\big(\mathcal{F}(\widehat{\boldsymbol{x}}_{l}|\boldsymbol{X}_{int},\boldsymbol{X}_{obs})\big) \\
    &\text{s.t.}&& \boldsymbol{B}_{d,l}^{\mathrm{tsp}}\odot \boldsymbol{M}_{d}(\boldsymbol{y},C_{l}^{\prime})\in\prod_{\mathbb{G}_{d,l,j}^{\mathrm{tsp}}}(\mathbb{S}_{+}),\quad l=1,\ldots,L  \\
    &&&\boldsymbol{B}_{d,l,j}^{\mathrm{tsp}}\odot\hat{f}_{j}(\widehat{\boldsymbol{x}}_{l})\in\prod_{\mathbb{G}_{d,l,j}^{\mathrm{tsp}}}(\mathbb{S}_{+}),\quad j\in J_{l},l=1,\ldots,L \\
    &&&\boldsymbol{B}_{d,l,j}^{\mathrm{tsp}}\odot\widehat{g}_{j}(\widehat{\boldsymbol{x}}_{l})\in\prod_{\mathbb{G}_{d,l,j}^{\mathrm{tsp}}}(\mathbb{S}_{+}),\quad j\in J_{l},l=1,\ldots,L \\
    &\text{var}&& \{\widehat{\boldsymbol{x}}_{l}\}{.} 
    \end{aligned}$}
    \label{e20}
\end{equation}

This problem can be viewed as a coupling of $L$ sub-problems, each constituting an SDP problem with fewer variables. Then different sub-problems will be jointly learned by different clients to solve the problem of insufficient computing resources, similar to Algorithm 2. 

\begin{figure}[t]
\centering
\includegraphics[width=0.95\columnwidth]{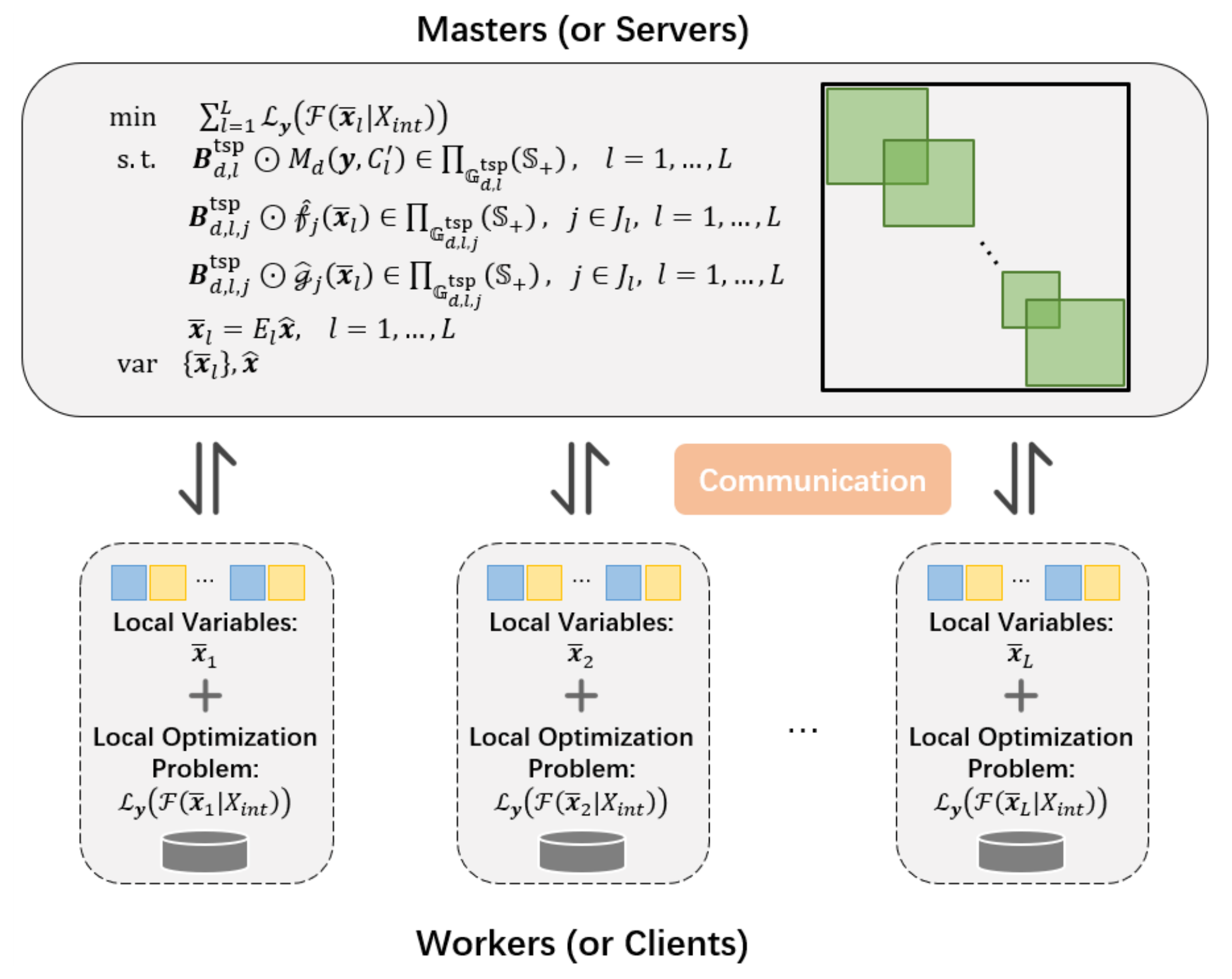} 
\caption{Distributed Architecture}
\label{fig2}
\vspace{-0.3cm}
\end{figure}

To ensure solution consistency, we add consensus variables and constraints for each sub-problem, reformulating it into a consensus problem as follows:
\begin{equation}
    \resizebox{.7\linewidth}{!}{$
            \displaystyle
    \begin{aligned}
    &\mathrm{min}&& \mathcal{L}_{y}\big(\mathcal{F}(\overline{\boldsymbol{x}}_{l}|\boldsymbol{X}_{int},\boldsymbol{X}_{obs})\big)  \\
    &\mathrm{s.t.}&& \boldsymbol{B}_{d,l}^{\mathrm{tsp}}\odot \boldsymbol{M}_{d}(\boldsymbol{y},C_{l}^{\prime})\in\prod_{\mathbb{G}_{d,l}^{\mathrm{tsp}}}(\mathbb{S}_{+}),  \\
    &&&\boldsymbol{B}_{d,l,j}^{\mathrm{tsp}}\odot\hat{f}_{j}(\overline{\boldsymbol{x}}_{l})\in\prod_{\mathbb{G}_{d,l,j}^{\mathrm{tsp}}}\left(\mathbb{S}_{+}\right),\quad j\in J_{l}, \\
    &&&\boldsymbol{B}_{d,l,j}^{\mathrm{tsp}}\odot\widehat{g}_{j}(\overline{\boldsymbol{x}}_{l})\in\prod_{\mathbb{G}_{d,l,j}^{\mathrm{tsp}}}\left(\mathbb{S}_{+}\right),\quad j\in J_{l}, \\
    &&&\overline{\boldsymbol{x}}_{l}=E_{l}\widehat{\boldsymbol{x}},\quad  \\
    &\mathrm{var}&& \{\overline{\boldsymbol{x}}_{l}\},\widehat{\boldsymbol{x}}{,} 
    \end{aligned}$}
    \label{e17}
\end{equation}
where $\overline{\boldsymbol{x}}_{l}$ are the local variables in $l_{th}$ worker, and $\widehat{\boldsymbol{x}}$ are the consensus variables in the master node. By formulating the consensus problem, it enables the design of distributed algorithm based on architecture such as parameter server. As shown in Fig. \ref{fig2}, communication is centralized around the server. Clients retrieve the consensus variable $\widehat{\boldsymbol{x}}$ from the server and transmit the local variable $\overline{\boldsymbol{x}}_{l}$ to it. By jointly solving all subproblems in Eq. (21), we will obtain a solution to the original problem.

\section{Experiments}

\subsection{Experiments with Different $\lambda_{obs}$}
\begin{figure}[!t]
\centering
\includegraphics[width=0.8\columnwidth]{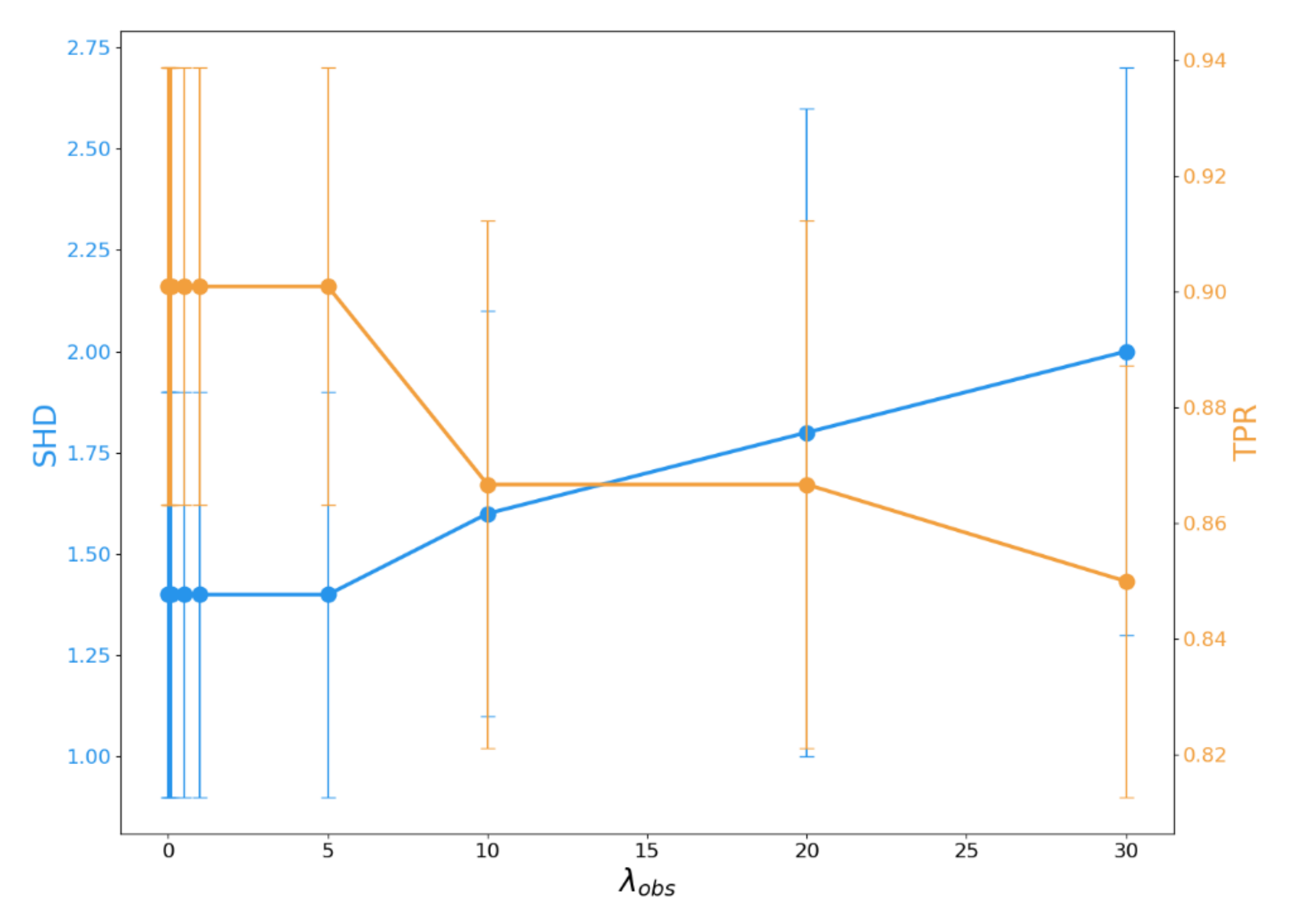}
\caption{Experiments with Different $\lambda_{obs}$.}
\label{sup_2}
\end{figure}

In the formulated bilevel optimization problem Eq. (11,12), we leverage both observational and interventional data to learn direction parameters of the causal structure, which aims to fully exploiting the causal information in datasets. Specifically, we introduce the weight coefficient $\lambda_{obs}$ to balance the influence of these two types of data. As mentioned earlier, interventions are the second level in the Pearl's causal ladder theory, which can more effectively reveal the causal relationship between variables. The experimental results are shown in the Fig. \ref{sup_2}. Analyzing the outcomes, it is evident that when $\lambda_{obs}$ is small, Bloom's performance remains consistently robust, effectively discerning causal structures. Conversely, a larger $\lambda_{obs}$ may lead to the identification of spurious causal relationships, which is primarily resulted by the dominance of observational data. Therefore, in the upper-level objective function, we preferentially opt for smaller $\lambda_{obs}$ values to enhance the discernibility of the causal structure. Furthermore, the experimental results substantiate that the introduction of interventional data indeed contributes to an improved accuracy in identifying the causal structure.

\subsection{Experiments with Different Iteration Times}
To analyze the sensitivity of the proposed method to the number of iterations, we tested the algorithm's performance under different iteration settings. Fig. \ref{fig_senexp}(a) shows the experiments with 5 nodes, and Fig. \ref{fig_senexp}(b) with 10 nodes. When the iteration count is relatively low, especially below 100, the performance is suboptimal. This is because the algorithm may not have sufficiently explored the solution space with a low iterations, leading to suboptimal performance. However, as the number of iterations increases, the algorithm's performance improves significantly. After reaching 300-500 iterations, the performance stabilizes. Specifically, the time required for the algorithm to run 500 iterations is around 29.3 seconds for 5 nodes and approximately 240.4 seconds for 10 nodes. Despite the increase in time required with more nodes, the overall convergence time remains acceptable. Based on the trade-off between algorithm performance and efficiency, the iteration number is generally set above 1000, the algorithm performance is relatively stable regarding this parameter setting. Hence, the algorithm is robust to the iteration parameter setting as long as it is set reasonably.

\subsection{Experiments on Large-scale Datasets}
\begin{figure*}[!t]
\centering
\includegraphics[width=1.85\columnwidth]{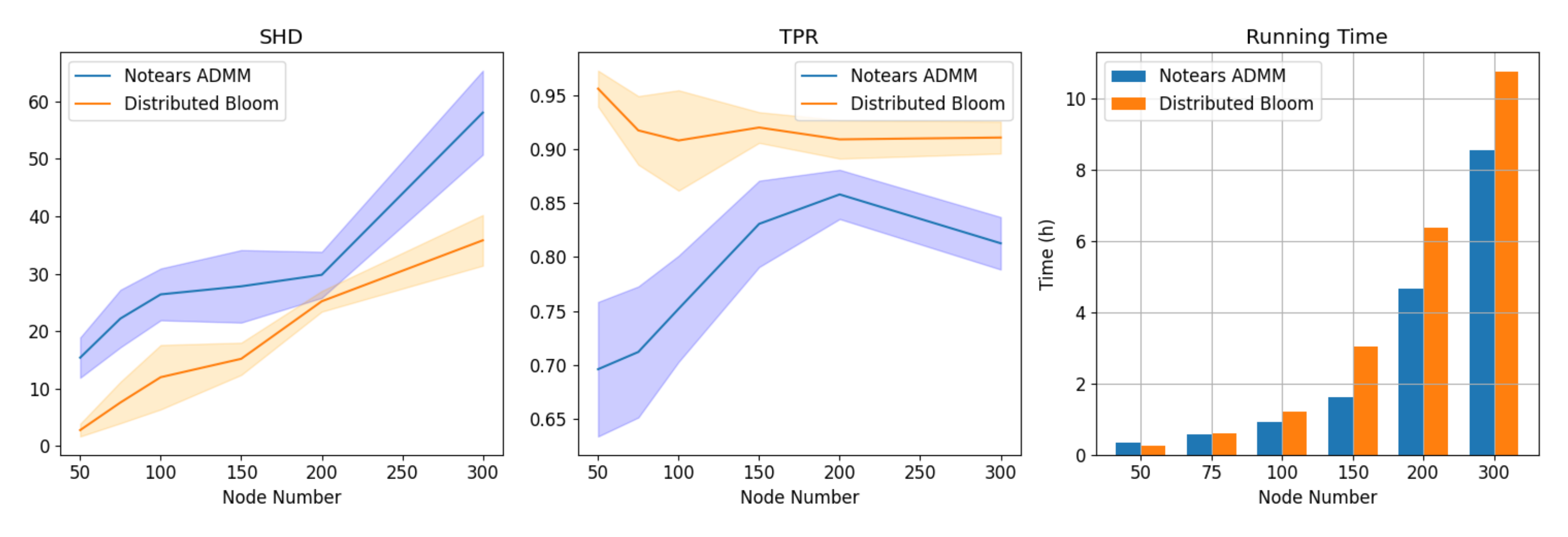}
\caption{Experiments on Large-scale Datasets.}
\label{morenodes}
\end{figure*}

\begin{figure}[!t]
    \centering
    \subfigure[5 Nodes]{
        \includegraphics[width=0.45\textwidth]{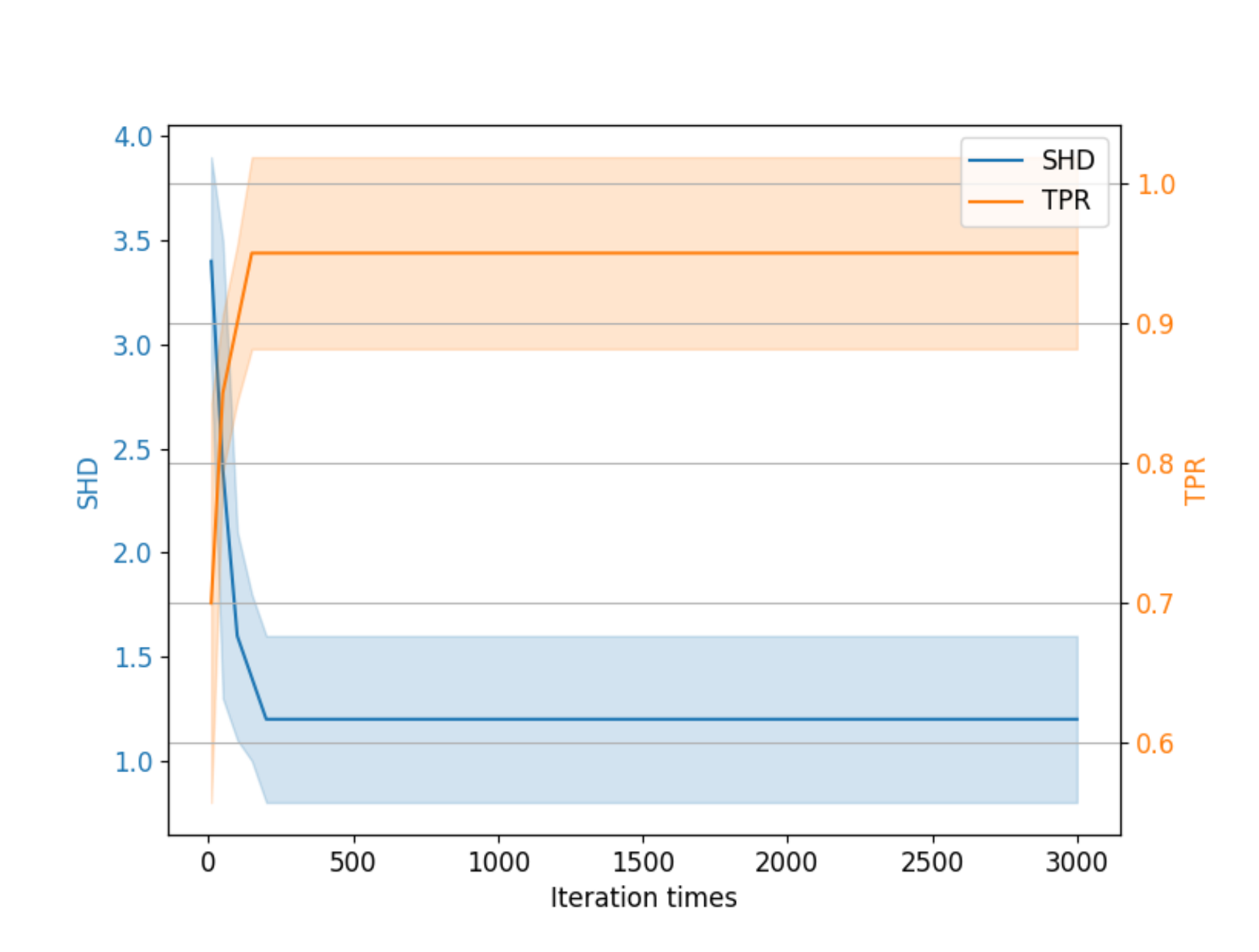}
    }
    \hspace{0.01\textwidth}
    \subfigure[10 Nodes]{
        \includegraphics[width=0.45\textwidth]{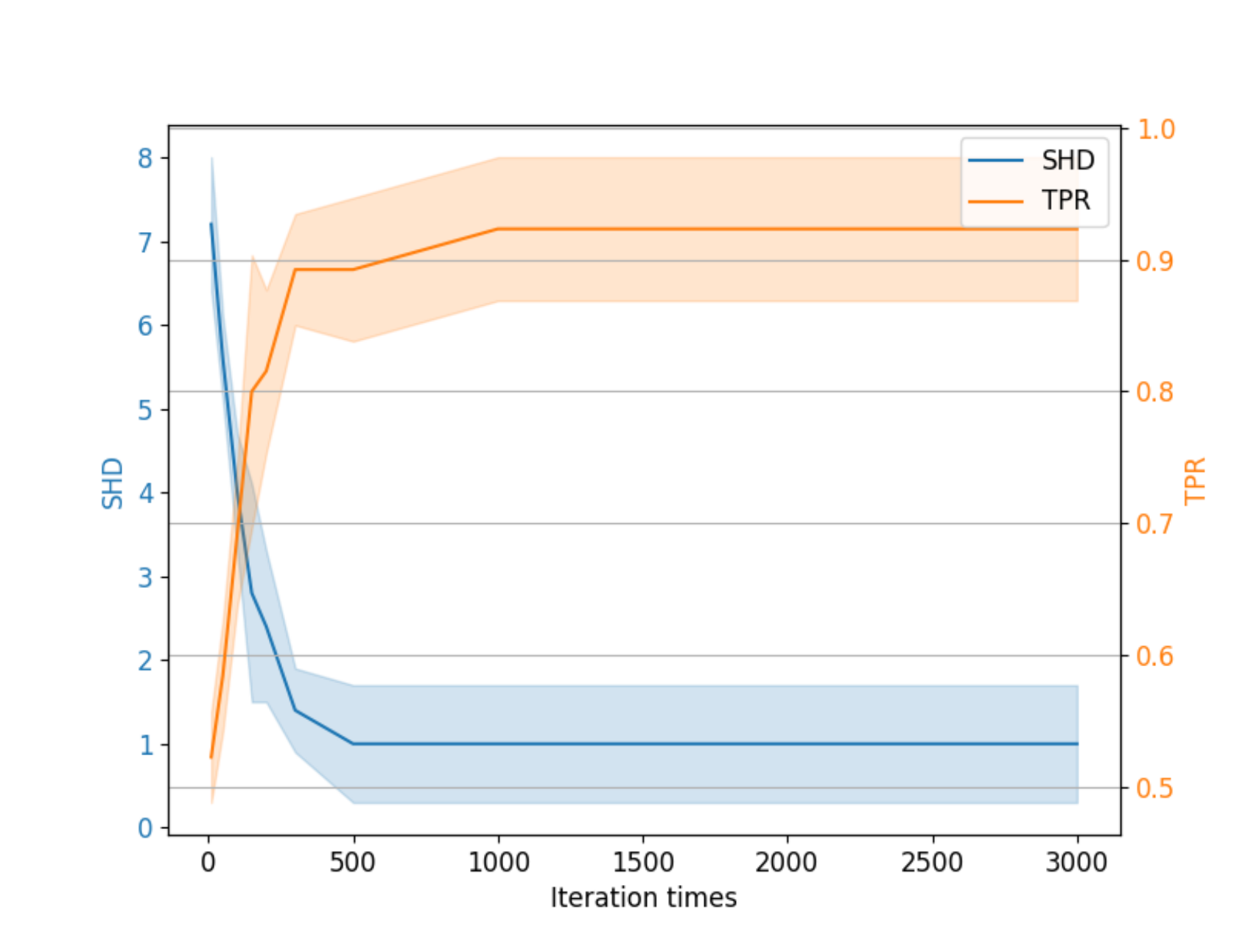}
    }
    \caption{Experiments with Different Iteration Times.}
    \label{fig_senexp}
\end{figure}

On large-scale datasets, we further compared the performance of the distributed Bloom algorithm and the federated learning-based Notears-ADMM algorithm \cite{ng2022towards}. The datasets are derived from a linear generative model, with a node-to-edge ratio of 1:1. Both algorithms were run for 1000 iterations, while the other parameters of Notears-ADMM are kept at their original settings. Fig. \ref{morenodes} show the SHD, TPR, and running time for both methods on all datasets. The experimental results indicate that as the number of nodes increases, the SHD for both methods also increases, but the SHD for distributed Bloom remains consistently lower than Notears-ADMM, demonstrating its superior performance for current problems. This is due to distributed Bloom’s use of interventional data, which effectively reduces spurious correlations. Additionally, the TPR for distributed Bloom remains consistently high, indicating its strong capability in identifying true causal relationships, partly attributed to the use of polynomial optimization. Regarding running time, although both methods are similar with fewer nodes ($\leq 100$), distributed Bloom takes longer as the number of nodes increases. This is because Bloom needs converting the original POP problem into an SDP relaxation problem, causing the running time to grow exponentially with the number of nodes. Therefore, in practical applications, a trade-off between running efficiency and accuracy might be necessary. Overall, distributed Bloom still demonstrates good applicability and performance on larger-scale datasets. Future research will focus on optimizing the efficiency of the proposed method. Many existing studies \cite{scalesdp} have explored how to accelerate scalable SDP problem, and we hope to apply these techniques to our algorithm.

\subsection{Experiments with Different Sampling Number}
\begin{figure}[!t]
\centering
\includegraphics[width=0.8\columnwidth]{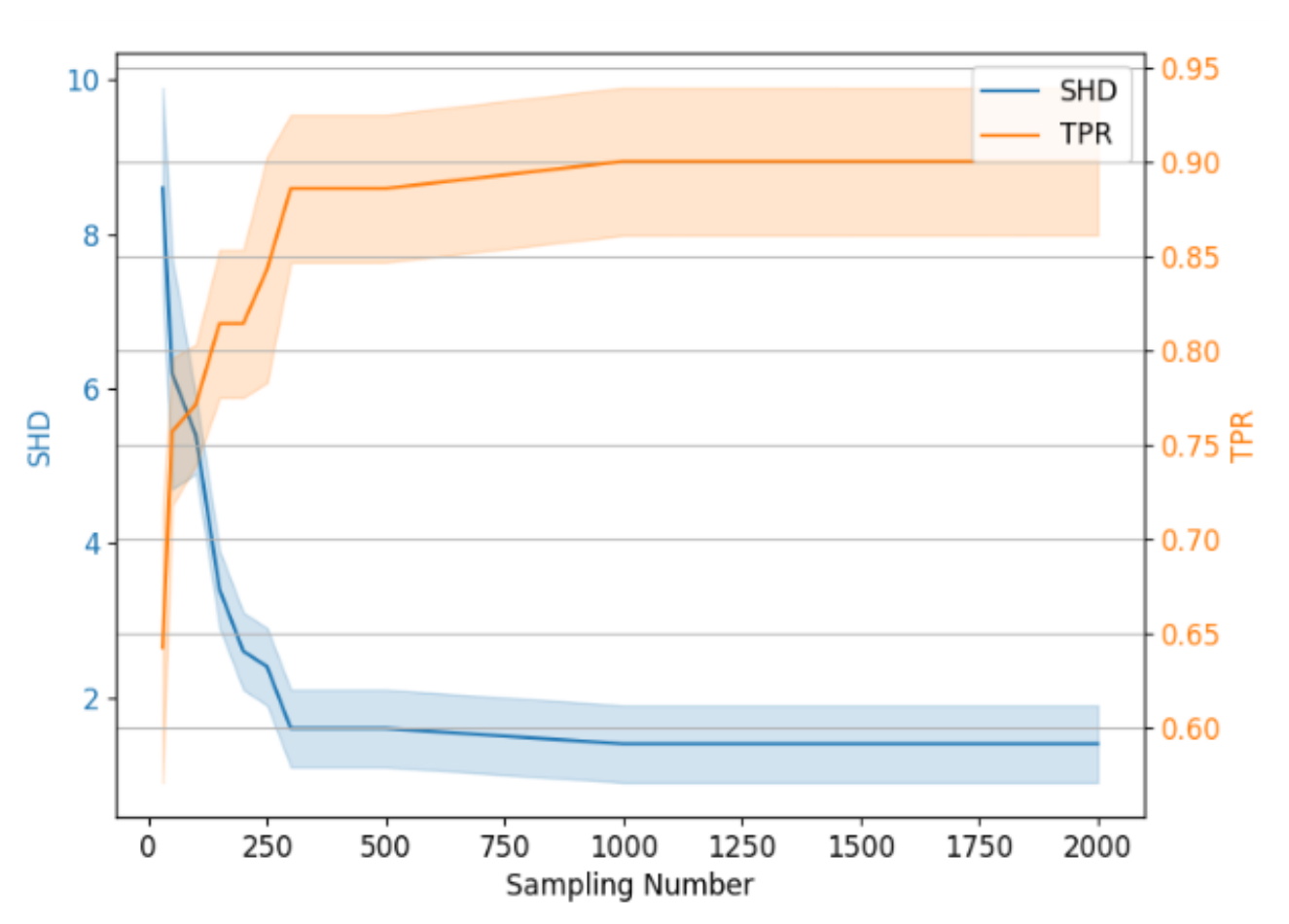}
\caption{Experiments with Different Sampling Number.}
\label{sup_1}
\end{figure}

We tested the performance of the proposed method in data sets with different sampling lengths (15 Nodes) in this experiment, and the results are shown in the Fig. \ref{sup_1}. The figures display the SHD and TPR of the algorithm under different settings. From the Fig. \ref{sup_1}, we can observe that as the sampling number increases, the performance of our method improves rapidly. Moreover, the proposed method reaches relatively high performance and stabilizes when the sample length is 300.

\begin{figure}[!t]
\centering
\includegraphics[width=0.8\columnwidth]{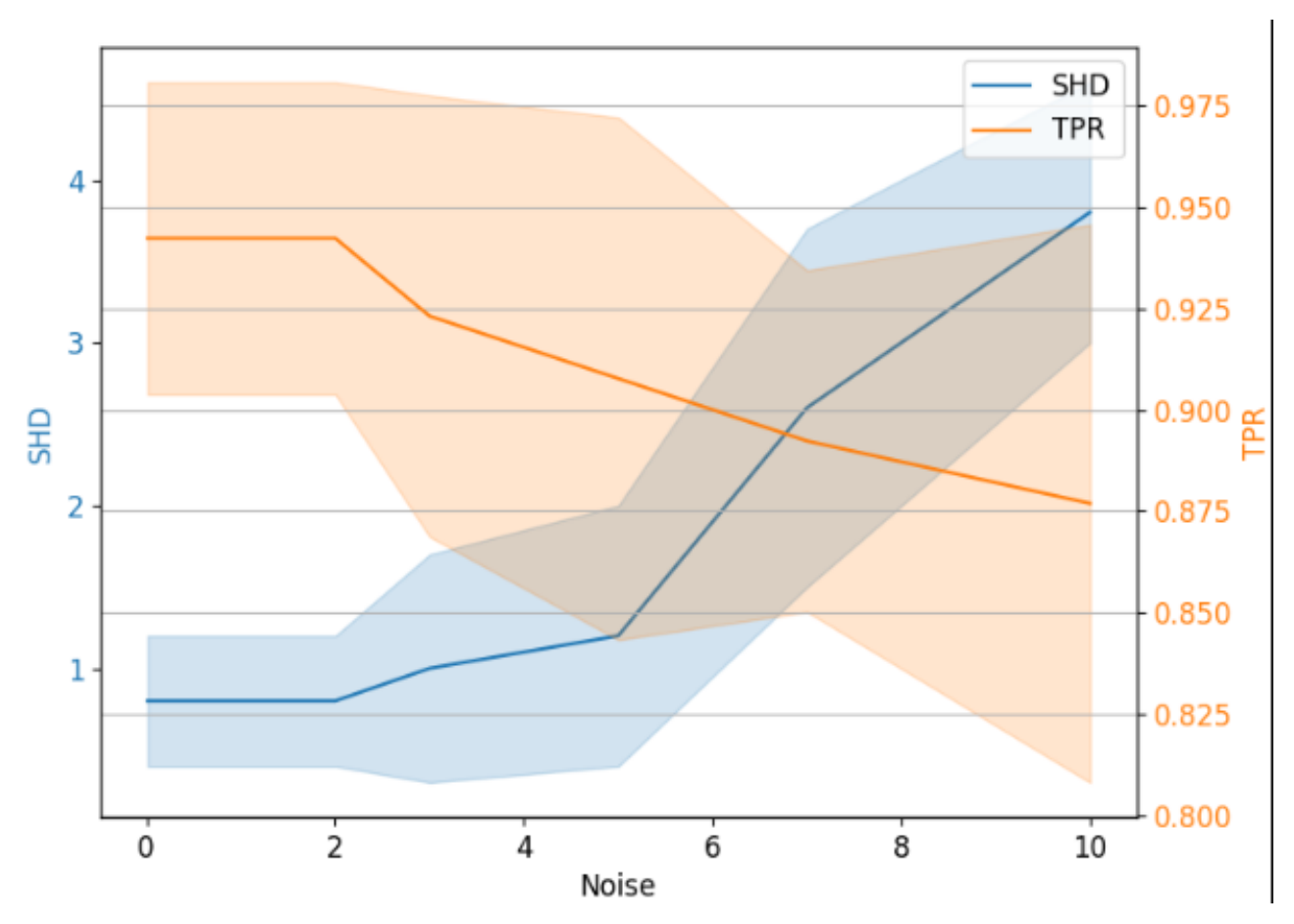}
\caption{Experiments with Different Noise.}
\label{sup_1_1}
\end{figure}

\subsection{Experiments with Different Noise}
In this study, we sampled data noise from Gaussian distributions with varying standard deviations (ranging from 0.1 to 10.0). The results of the experiment are depicted in Fig. \ref{sup_1_1}, with the vertical axis indicating the standard deviation of the noise. We report the algorithm's SHD and TPR. The figure shows that as the noise level increases, our algorithm maintains relatively high performance until the standard deviation reaches to 6.0. This partly illustrates the stability of the proposed algorithm in high-noise conditions.

\subsection{Convergence Experiment}
Bloom initially relaxes the formulated bilevel polynomial problem and learn the causal structure by solving a sequence of SDP problems. Different from existing continuous optimization methods for causal discovery, we utilize Interior Point Method (IPM) and its variants as the solver for our problem. IPM is a powerful optimization algorithm that progresses towards the global optimum in each iteration, typically resulting in a faster global convergence rate. In contrast, existing continuous optimization methods often rely on gradient descent, updating parameters along the negative gradient direction of the objective function to gradually approach the optimal solution. In some cases, especially for non-convex problems, gradient descent may get trapped in local optima, leading to slower convergence rates. Additionally, IPM essentially employs Newton method for unconstrained convex subproblems. Compared to the gradient descent-based works, which shows a sub-linear or linear convergence rate, Newton method theoretically offers a quadratic convergence rate. Thus, our method achieves faster convergence

\begin{figure}[!t]
\centering
\includegraphics[width=0.9\columnwidth]{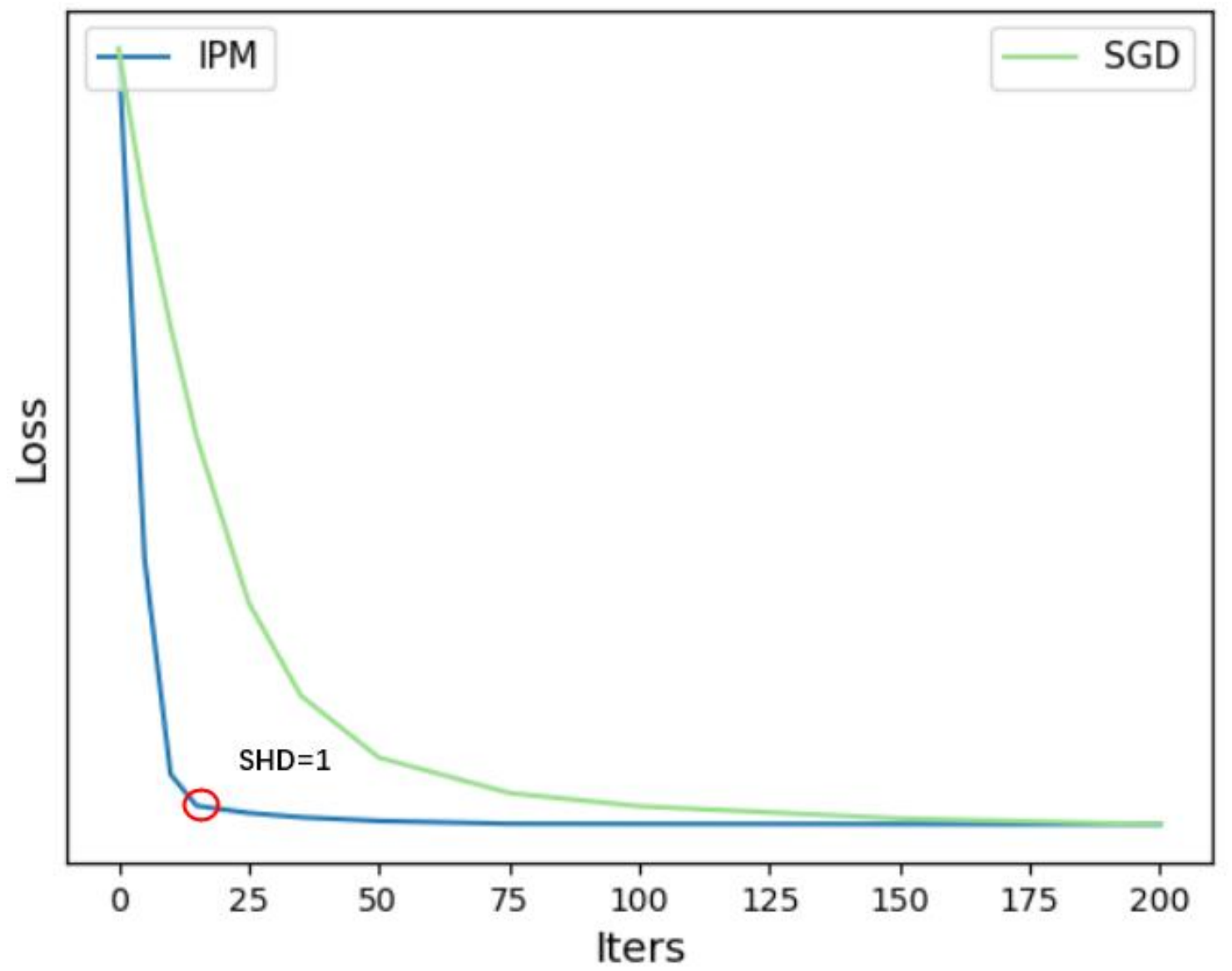}
\caption{Algorithm Convergence.}
\label{sup_3}
\end{figure}

To verify this idea, we conducted additional experiments to analyze the convergence and speed of the algorithm. A comparison was made between the convergence of the SGD-based causal discovery method and the proposed algorithm on a data set (nodes = 5, edges = 8), as illustrated in the Fig. \ref{sup_3}. The figure indicates that our method can achieve better experimental results in approximately 20 iterations, with the SHD of the learned causal graph being 1. In contrast, the SGD-based method requires more iterations to achieve convergence. This is due to the use of a global convergence approach (i.e., IPM) used to solve the SDP problems, which essentially utilizes the Newton method, known for its quadratic convergence, while the gradient descent method typically converges linearly. Furthermore, the experimental results presented in the Section 4 also demonstrate the superior performance of our proposed method.

\end{document}